\documentclass{article}

\usepackage{microtype}
\usepackage{graphicx}
\usepackage{booktabs} 

\usepackage{url}
\usepackage{breakurl}
\usepackage[breaklinks]{hyperref}

\usepackage[accepted]{sysml2019}

\usepackage{listings}
\usepackage[utf8]{inputenc}
\usepackage{amsthm}
\usepackage{amssymb}
\usepackage{todo}
\usepackage{tikz}
\usepackage{subfig}
\usepackage{media9}

\sysmltitlerunning{RL Metaoptimization on a Distributed System}

\begin{document}

\twocolumn[
\sysmltitle{Metaoptimization on a Distributed System for Deep Reinforcement Learning}




\begin{sysmlauthorlist}
\sysmlauthor{Greg Heinrich}{aff}
\sysmlauthor{Iuri Frosio}{aff}
\end{sysmlauthorlist}

\sysmlaffiliation{aff}{NVIDIA, Santa Clara, USA}

\sysmlcorrespondingauthor{Greg Heinrich}{gheinrich@nvidia.com}
\sysmlcorrespondingauthor{Iuri Frosio}{ifrosio@nvidia.com}


\vskip 0.3in
\begin{abstract}
Training intelligent agents through reinforcement learning is a notoriously unstable procedure.
Massive parallelization on GPUs and distributed systems has been exploited to generate a large amount of training experiences and consequently reduce instabilities, but the success of training remains strongly influenced by the choice of the hyperparameters.
To overcome this issue, we introduce HyperTrick, a new metaoptimization algorithm, and show its effective application to tune hyperparameters in the case of deep reinforcement learning, while learning to play different Atari games on a distributed system.
Our analysis provides evidence of the interaction between the identification of the optimal hyperparameters and the learned policy, that is typical of the case of metaoptimization for deep reinforcement learning.
When compared with state-of-the-art metaoptimization algorithms, HyperTrick is characterized by a simpler implementation and it allows learning similar policies, while making a more effective use of the computational resources in a distributed system.
\end{abstract}
]



\printAffiliationsAndNotice{}  

\section{Introduction}
\label{sec:Introduction}

The Reinforcement Learning (RL) field has been recently revitalized by the advent of Deep Learning (DL), with the development of new training algorithms~\cite{Lil15, mnih-dqn-2015, Mni16} effectively applied in several fields including, among others, gaming, robotics, and finance.
Despite these recent successes, training in RL remains an unstable procedure that requires fine hyperparameter tuning. 

The causes of instability in RL are copious.
High correlation between training data generates instability; the fact that it can be reduced by collecting data from a wide set of agents, acting in parallel but in different environments~\cite{Mni16, Bab16, Bab17}, has triggered the investigation of GPUs and distributed systems for the simulation of RL environments~\cite{NairSBAFMPSBPLM15, Esp18, Sto18}.
Value-based algorithms assign a numerical value to each state observed by the agent, and develop policies aimed at reaching high value states; training instabilities are associated in this case with Monte Carlo sampling, that requires to play an entire episode before computing an unbiased, high variance estimate of the value of each state.
N-steps methods reduce the variance by boosting~\cite{Lil15, Mni16}, at the cost of increasing the bias, whereas identifying the optimal bias-variance trade-off is a non-trivial problem~\cite{Buc18}.
Instability is also caused by falls into local minima associated to sub-optimal policies, that can be prevented by adding an entropy term to the cost function which favours exploration over exploitation ~\cite{Lil15, mnih-dqn-2015, Mni16}.
Beyond this, far- or near-sighted agents can be more or less prone to fall into local minima when learning different tasks.

Modern RL algorithms, and especially those based on DL, use several hyperparameters to control most of the previously mentioned instability factors.
Careful hyperparameter tuning is required to balance between speed, effectiveness, and stability of the training process.
The general problem of identifying an optimal set of hyperparameters, while solving the underneath optimization problem, is referred to as \emph{metaoptimization}. In RL it is often solved only after a long sequence of ineffective, time-consuming, trial-and-error, training attempts.

We tackle metaoptimization for RL on a distributed system, to achieve convergence towards a reasonable, learned policy with a minimal effort for the user.
When compared to traditional metaoptimization, and in particular when considering a distributed system, the specific case of deep RL shows some peculiarities, like the fact the the choice of the hyperparameters may strongly affect the computational cost of an experiment, and the high instability of the training procedure.
Our contribution here is a thorough analysis of metaoptimization for deep RL and a new algorithm particularly suitable for this case.
More in detail:
\begin{itemize}
\item We introduce HyperTrick, a metaoptimization procedure that generalizes Successive Halving~\cite{Jam16}.
HyperTrick is implemented on top of MagLev~\cite{Fa18}, a recently introduced training and inference framework to manage distributed systems.
\item We demonstrate the effectiveness of HyperTrick in deep RL, to learn policies for several Atari games through GA3C~\cite{Bab16,Bab17}, with minimal effort on the user side for hyperparameter setting.
\item We show evidence of the interaction between the optimized set of hyperparameters and the learned policies, which is peculiar of metaoptimization for RL.
\item We compare HyperTrick with the state-of-art Hyperband~\cite{Li16} and show that HyperTrick has a simpler implementation that does not require any support for preemption, it achieves a higher occupancy by effectively releasing and reallocating computational resources during metaoptimization, while reaching similar results in terms of learned policies. 
\end{itemize}
\section{Related work}
\label{sec:RelatedWork}

\paragraph{RL Algorithms:} Recent advances have been triggered by the development of novel algorithms for DL agents, but such advances did not come for free.
RL is notoriously unstable when the action-value function is estimated by a nonlinear function approximator such as a Deep Neural Network (DNN), because of correlations in the sequence of observations, changes in the policy causing changes in the data distribution during training, and correlations between the action-values and the target values~\cite{mnih-dqn-2015}.
The DQN approach~\cite{mnih-dqn-2015} reduces these instabilities through a large replay memory and an iterative update rule that adjusts the action-values towards target values that are only periodically updated.
Other learning procedures inspired by DQN achieve faster and more stable convergence: Prioritized DQN~\cite{SchaulQAS15} gives priority to significant experiences in the replay memory.
Double-DQN~\cite{HasseltGS15} separates the value function estimation and action selection, reducing the DQN tendency to be overly optimistic when evaluating its choices.
Dueling Double DQN~\cite{WangFL15} goes a step further by explicitly splitting the computation of the value and advantage functions within the network.

Actor-critic methods, like A3C~\cite{Mni16} and its GPU version, GA3C~\cite{Bab16,Bab17}, outperform the DQN methods.
Actor-critic methods alternate policy evaluation and improvement steps; both the actor and the critic are trained during learning.
The critic is often modelled by a n-step bootstrapping method, which reduces the variance and stabilizes learning when compared to pure policy gradient methods.
Parallelization allows simulating multiple environments: it increases the diversity of the experiences collected from the agents, reducing the correlation in the observations.
An entropy term is also generally included to favour exploration.
Nonetheless, the stability and convergence speed of DQN and actor-critic methods strongly depend on the choice of several hyperparameters, such as the learning rate, or the number of steps used for bootstrapping.
For instance, 50 different learning rates are tested for each Atari game in~\cite{Mni16}, to guarantee convergence towards a reasonable policy.

\paragraph{RL on Distributed Systems:} Distributed systems are commonly used in RL, with the aim of generating as many experiences as possible, as convergence towards an optimal policy is achieved only if a sufficiently large number of experiences is consumed by the RL agent.
For instance, Gorilla DQN~\cite{NairSBAFMPSBPLM15} outperforms DQN by using $100$ actors on $31$ machines, $100$ learners, and a central parameter server with the DNN model.
IMPALA~\cite{Esp18} employs hundreds of  CPUs and it solves an Atari game in a few minutes; a similar result can be achieved by resorting to a multi-GPU system~\cite{Sto18}.
Beyond speeding up RL, the aforementioned approaches achieve training stability by dramatically increasing the number of simulated environments, and thus increasing the batch size and the diversity of the collected experiences.
Nonetheless, even in this case hyperparameter setting remains critical to guarantee convergence, and the proper configuration has to be identified through a time consuming trial-and-error procedure.
For instance, it has been shown that the learning rate and the batch size have to be properly scaled to guarantee the convergence of many deep RL algorithms on large distributed systems, but it has also been observed that some of them (like Rainbow-DQN) do not scale beyond a certain point~\cite{Sto18}.
Moreover, not all RL algorithms naturally scale to a distributed implementation: for instance GA3C~\cite{Bab16,Bab17} can hardly benefit from distribution on a multi-GPU system, as it is limited by the CPU time required to generate experiences and bandwidth to move data from the RAM to the GPU.
On the other hand, we can still leverage a distributed system to explore the hyperparameter space and learn an optimal policy; this is the metaoptimization approach described here.

\paragraph{Metaoptimization on Distributed Systems:}
Metaoptimization consists in finding a set of optimal hyperparameters, while solving an underneath optimization problem that depends on such hyperparameters.
Grid and random search are basic metaoptimization methods, based on \emph{parallel search}: a wide exploration of the hyperparameter space is performed by parallel optimization processes with different hyperparameters; the best hyperparameter set is consequently identified, but one limit of this approach is that the optimization processes do not share any information.
Other basic metaoptimization procedures, such as hand tuning or Bayesian-Optimization~\cite{Sha16}, follow a \emph{sequential search} paradigm, where the results achieved by completed optimization processes drive the selection of new hyperparameters.
The search for the optimal hyperparameter set is in this case local, and evidence of the optimal setting generally emerges only after a large number of evaluations - this is a limiting factor, especially when the underneath optimization problem is computationally intensive.
When the underneath problem is solved iteratively, partial results can be used as proxy and hyperparameter configurations that are deemed less promising can be abandoned quickly; this scheme is referred to as \emph{Early Stopping}.


Population Based Training (PBT,~\cite{Jad17}) leverages the benefits of parallel search, sequential search, and early stopping, merging them into a single, metaoptimization procedure that automatically selects hyperparameters during training, while also allowing online adaptation of the hyperparameters to enable non-stationary training regimes and the discovery of complex hyperparameter schedules; it performs online model selection to maximize the time spent on promising models.
PBT is naturally implemented on a distributed system, by assigning one ore more optimization processes to each node.
Hyperband~\cite{Li16}, an extension of Successive Halving~\cite{Jam16}, is another algorithm that uses adaptive resource allocation and early stopping to solve metaoptimization problems.
In Successive Halving, the exploration of the hyperparameter space is performed in multiple phases.
Each phase is given a total resource budget $B$, equally divided between $N$ workers, where each worker is solving the underneath optimization problem using a different set of hyperparameters.
The worst half of the workers are terminated at the end of each phase, while the other ones are allowed to run.
The main issue of this approach is that, for a fixed $B$, it is not clear a priori whether it is better to consider many hyperparameter configurations (large $N$) with a small average training time, or a small number of configurations $N$ with longer average training times.
Hyperband addresses the problem of balancing breadth (large $N$) versus depth (small $N$) search by calling Successive Halving as a subroutine and considering several possible values of $N$ for a fixed $B$; each $(B, N)$ pair is called a \emph{bracket} in Hyperband.
Hyperband's inputs are the maximum amount of resource allocated to a single configuration, $R$; and the proportion of configurations discarded in each round of Successive Halving, $\eta$.
In the first bracket, Hyperband sets $N$ to its smallest value for maximum exploration and runs Successive Halving for the given $B/N$ ratio (under the constraint imposed by $R$).
For any successive bracket, Hyperband reduces $N$ by a factor of approximately $\eta$ until, in the last bracket, one final configuration (performing classical random search) is left.
In practice, Hyperband performs a grid search over $N$ by running several instances of Successive Halving.
Under this scheme, slow learners who may initially under-perform are given a chance to run for longer and may ultimately yield better results.

Unfortunately, any attempt to implement Successive Halving and Hyperband on a distributed system easily reveals some practical issues.
These algorithms are well suited for systems with a unique node or a set of equivalent nodes, but in the case of an heterogeneous distributed system, the effective assignment of a fixed budget $B/N$ to any worker is problematic: nodes associated to fast workers may be idle, waiting to synchronize with the slow nodes at the end of each phase.
Quite remarkably, this may happen even on a homogeneous system if the hyperparameters affect the computational cost of the underneath optimization problem, like in the case of metaoptimization for RL.
Although an asynchronous variation of Successive Halving can be used to partially solve this problem~\cite{Li18}, a second issue is that, if the number of compute nodes is not large enough, some workers need to yield to other workers at the end of each phase and resume execution.
This requires explicit support for preemption, and introduces an additional overhead for context switching.
These two issues may be alleviated by allowing mapping of one worker to multiple nodes over the course of the training process, which however requires again a non trivial implementation and incurs a context switching cost.

\section{HyperTrick on MagLev}
\label{sec:metaOptimization}

We propose a metaoptimization procedure based on parallel search and early stopping, which frees computational resources during the process, re-allocates them for new experiments, and does not require support for preemption.
We perform our experiments on the MagLev platform, whose architecture is briefly described in the following.

\subsection{The MagLev Architecture}
\label{sub-maglev}

\begin{figure}[t]
\centering
\includegraphics[width=0.48\textwidth,clip=true,trim=0.2cm 1.1cm 0.2cm 0.2cm]{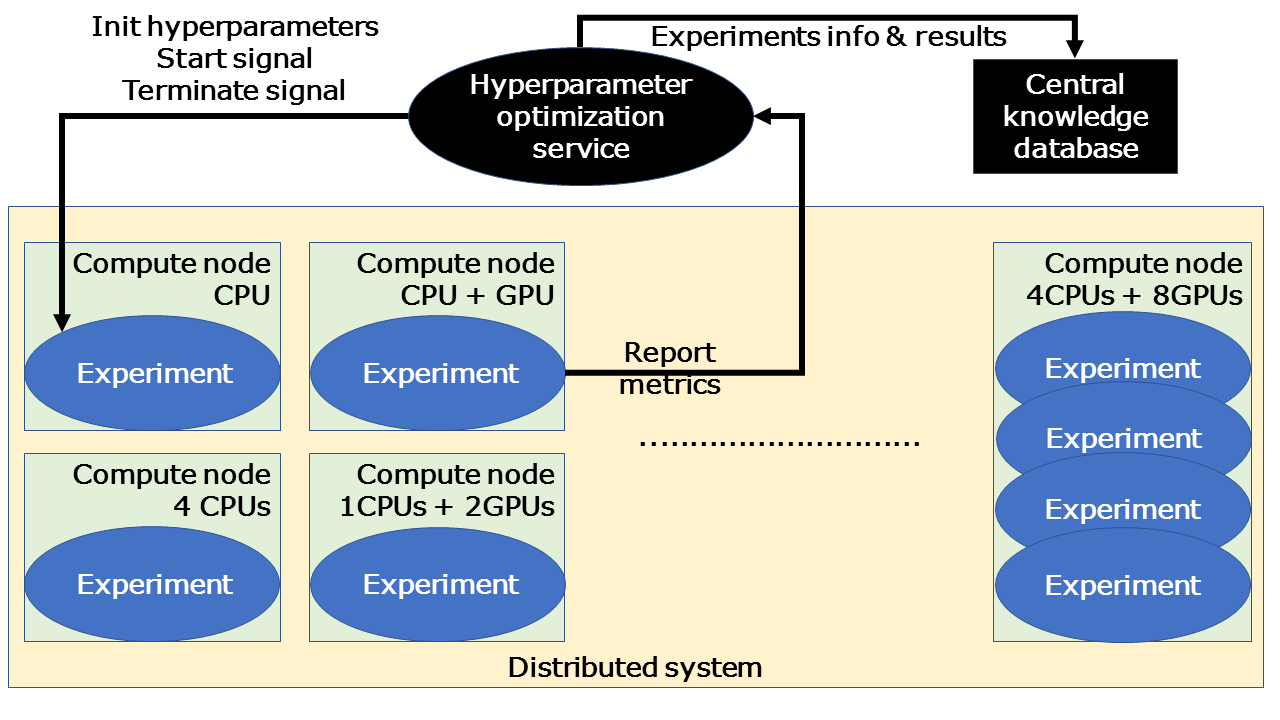}
\caption{The architecture of the MagLev platform, supporting heterogeneous nodes with CPUs and GPUs. The hyperparameter optimization service randomly initializes the hyperparameters of each experiment, and collects metrics and results into a central knowledge database. Each node runs one or more experiments at the same time. The optimization service can also terminate some of the experiments to re-allocate the compute nodes.}
\label{fig:maglev}
\end{figure}

MagLev is a platform built over Kubernetes ~\cite{Kub} that supports the execution of parallel experiments on a distributed system, with homogeneous or heterogenous nodes, each including one or multiple CPUs and GPUs.
Metaoptimization can be implemented in different flavors in MagLev by performing a number of parallel optimization experiments, each exploring one point of the hyperparameter space.
A hyperparameter optimization service runs in MagLev to this aim (see Fig.\ref{fig:maglev}).
Each experiment is executed by a \emph{worker}; each compute node can host one or multiple workers (and thus run multiple experiments) at the same time.
MagLev allows each experiment to continuously expose information about its status, as well as other metrics.
Typically each worker executes an experiment in multiple \emph{phases} and reports a set of metrics to the hyperparameter optimization service at the end of each phase.
The hyperparameter optimization service also manages the initial sampling of the hyperparameter space and it is backed by a central knowledge database that collects information about experiments, their hyperparameter configurations, and the reported metrics.
The workers periodically query the service to be notified whether to continue running or not.
This allows implementing (among other procedures) several metaoptimization algorithms that release and re-allocate computational resources, including our metaoptimization method, illustrated in Section \ref{sub-hypertrick}. 



\subsection{HyperTrick}
\label{sub-hypertrick}

We propose HyperTrick, a metaoptimization algorithm partially inspired by Hyperband~\cite{Li16}, Successive Halving~\cite{Jam16}, and PBT~\cite{Jad17}, aimed at improving the utilization of parallel resources in a distributed system, especially when the hyperparameter configuration affects the computational cost of an experiment, by merely managing the early termination of unpromising workers.
We describe here our asynchronous, multithread implementation, which suits well on a distributed system, although a single-thread implementation is also possible.

In HyperTrick, each worker explores one hyperparameter set over a number of phases $N_p$, where each phase may correspond to a number of training iterations, a given amount of wall-clock time, or any other user-defined arbitrary units of work.
Beyond $N_p$, HyperTrick's inputs are the initial number of workers, $W_0$, and the target eviction rate $r$, which is the expected ratio of workers terminated after each phase, although HyperTrick stochastically allows a different ratio of workers to proceed to the next phase.
The first step in HyperTrick is to launch a number of experiments equal to the minimum between $W_0$ and the number of nodes in the distributed system, $N$ (lines 2-3 in Algorithm~\ref{alg:hypertrick}). 
Differently from Successive Halving and Hyperband, HyperTrick does not employ any synchronization mechanism: each worker runs independently from others.
While different workers are running (line 4 in Algorithm~\ref{alg:hypertrick}), they may be in different phases.
Workers asynchronously report performance metrics at the end of each phase to the central hyperparameter optimization service (line 5), which stores the statistics (line 6) and then uses the HyperTrick's rule to decide whether to let a worker continue (lines 7).
When a worker is terminated, the compute node is reallocated to a new worker to investigate a new set of hyperparameters, starting from the first phase (lines 8-10 in Algorithm~\ref{alg:hypertrick}).
HyperTrick eventually returns the best observed configuration (line 11).
\renewcommand{\lstlistingname}{Algorithm}
\begin{figure*}[th]
\begin{center}
\definecolor{mygreen}{rgb}{0,0.6,0} 
\definecolor{mygray}{rgb}{0.95,0.95,0.95} 
\definecolor{mymauve}{rgb}{0.58,0,0.82}
\begin{lstlisting}[label={alg:hypertrick}, basicstyle=\ttfamily\scriptsize, breaklines=true, numbers=right, language=C++, tabsize=1, frame=single, stringstyle=\color{mymauve}, commentstyle=\color{mygreen}, keywordstyle=\color{blue}, backgroundcolor=\color{mygray}, caption = HyperTrick.]
HyperTrick(W_0, r, N_p, N) //W_0 workers, r target eviction rate, N_p phases, N computational nodes.
    for (i=0; i<min(W_0, N); ++i)
        launch_experiment_thread(i, r, N_p)   // launch experiment on i-th node with random hyperparameters
    while(no_more_experiment_running() == false)
        [n, m, p, h] = wait_for_experiment_report() // wait for one experiment with hyperparameters h reporting metric m at phase p on node n
        stats = store_statistics(stats, m, p, h) // store metrics m at phase p and hyperparameters h in stats
        if (terminate_experiment_thread_if_needed(n, m, p, stats) == true) // kill experiment with HyperTrick rule
            if (i < W_0 - 1)
                ++i
                launch_experiment_thread(n, r, N_p) // launch experiment on n-th node with random hyperparameters
     return best_of(stats)
\end{lstlisting}
\end{center}
\end{figure*}

The HyperTrick's rule to decide whether a worker should continue or not at the end of each phase is the following.
Within each phase, HyperTrick first operates in Data Collection Mode (DCM), collecting metrics and letting all workers proceed to the next phase.
Once sufficient statistics have been collected, HyperTrick switches to the Worker Selection Mode (WSM) for that phase and starts terminating under-performing workers.
Let $W_p$ be the number of workers at phase $p$. For the given target eviction rate $r$, the expected value of $W_p$ is given by:
\begin{equation}
E[W_p] = W_0 (1-r)^p.
\label{eq:n_p}
\end{equation}
The number of workers required to complete the phase $p$ before switching from DCM to WSM, $W_p^{DCM}$, is:
\begin{equation}
W_p^{DCM}=W_0(1-\sqrt{r})(1-r)^p.
\label{eq:n_DCM}
\end{equation}
Once HyperTrick switches to WSM in phase $p$, any worker $w$ that reports a metric $m_w(p)$ in the lower $\sqrt{r}$ quantile is terminated.
We demonstrate in the following that using Eq. (\ref{eq:n_DCM}) to switch from DCM to WSM leads to the expected target eviction rate in Eq.  (\ref{eq:n_p}).
\begin{proof}
Assumption: the process $m_w(p)$, which returns the metric of a worker $w$ for phase $p$, is stationary.

Base case $p=0$: For $p=0$, the total number of workers is always the initial number of workers $W_0$; Eq. (\ref{eq:n_p}) predicts $E[W_0] = W_0(1-r)^0 = W_0$, thus it holds for $p=0$.

Inductive hypothesis: Suppose Eq. (\ref{eq:n_p}) holds for all values of $p$ up to some $k$, $k \geq 0$; then, at the beginning of phase $k$, the expected total number of workers is $E[W_k] = W_0(1-r)^k$.

Inductive step: in phase $k$, the first $W_k^{DCM}$ workers are allowed to continue unconditionally, whereas the remaining $W_k^{WSM}$ workers can be terminated by HyperTrick in WSM. We have:
\begin{eqnarray}
E[W_k^{WSM}] = E[W_k - W_k^{DCM}] = \nonumber \\
W_0(1-r)^k - W_0(1-\sqrt{r})(1-r)^k = \nonumber\\
W_0\sqrt{r}(1-r)^k.
\end{eqnarray}
Out of the $W_k^{WSM}$ workers, those that report $m_w(k)$ in the lower $\sqrt{r}$ quantile are terminated.
Because $m_w$ is stationary, the probability of a worker being terminated is $\sqrt{r}$.
If $W_k^{T}$ is the number of workers to terminate then:
\begin{equation}
E[W_k^{T}] = E[\sqrt{r} W_k^{WSM}] = rW_0(1-r)^k.
\end{equation}
The expected number of workers at the beginning of the next phase $k+1$ is then equal to:
\begin{eqnarray}
E[W_{k+1}] = E[W_k - W_k^{T}] = \nonumber  E[W_k] - E[W_k^{T}] = \nonumber \\ 
W_0(1-r)^k - rW_0(1-r)^k = W_0(1-r)^{k +1}
\end{eqnarray}
Therefore Eq. (\ref{eq:n_p}) holds for $p=k+1$.
By induction, it also holds for all $n \in \mathbb{N}$.
\end{proof}

There are several reasons for different workers to reach the end of a phase in different times: a worker may have been scheduled early or late, running on a fast or slow node, or assigned a more (or less) computationally efficient hyperparameter set.
HyperTrick favors early or fast workers in the selection and balances in this way between breadth versus depth search; it takes advantage of the unpredictability of worker scheduling, run time, and reported metrics, to give early workers a higher chance to continue and let them increase the depth of their search, while late workers are discouraged.
Across successive phases, HyperTrick requires low performers to be increasingly early, and laggards to perform increasingly well.
Since it does not synchronize workers and immediately re-allocates any idle node to a new worker, HyperTrick also achieves an effective utilization of the available computational resources.
This is a main advantage of HyperTrick when compared to Successive Halving and Hyperband, whose implicit synchronization mechanism at the end of each phase forces the fast workers to wait for the slow ones.
Idle nodes can be avoided in Successive Halving and Hyperband by allowing a dynamic scheduling of the workers on the nodes, but at the cost of implementing a preemption and yielding mechanism; on the other hand, HyperTrick does not require any preemption management, and does not incur any additional cost associated with context switches.
A beneficial side effect of HyperTrick is that, even in the case of occasional failures of the workers, the effect is local to the worker that experienced the failure. The experiment that was running on the worker may be retried, or ignored, without affecting other workers.
The price to be paid is the introduction of a potential bias in the final metaoptimization result, due to the random advantage given to the workers that are scheduled early.
Our experimental results show that, at least in the case of metaoptimization for RL considered here, this is of little importance in practice.

\newcounter{wavenum}

\setlength{\unitlength}{0.75cm}
\newcommand*{\clki}{
  \draw (t_cur) -- ++(0,.3) -- ++(.5,0) -- ++(0,-.6) -- ++(.5,0) -- ++(0,.3)
    node[time] (t_cur) {};
}

\newcommand*{\bitvector}[3]{
  \draw[fill=#3] (t_cur) -- ++( .1, .3) -- ++(#2-.2,0) -- ++(.1, -.3)
                         -- ++(-.1,-.3) -- ++(.2-#2,0) -- cycle;
  \path (t_cur) -- node[anchor=mid] {#1} ++(#2,0) node[time] (t_cur) {};
}

\newcommand*{\firstphase}[2]{
    \bitvector{#1}{#2}{green!50}
}

\newcommand*{\secondphase}[2]{
    \bitvector{#1}{#2}{blue!50}
}

\newcommand*{\thirdphase}[2]{
    \bitvector{#1}{#2}{orange!50}
}

\newcommand*{\fourthphase}[2]{
    \bitvector{#1}{#2}{red!50}
}

\newcommand*{\pending}[1]{
  \draw[ultra thick,black!50] (t_cur) -- ++(#1,0) node[time] (t_cur) {};
}

\newcommand*{\completed}[1]{
  \draw[ultra thick,green!50] (t_cur) -- ++(#1,0) node[time] (t_cur) {};
}

\newcommand*{\killed}[1]{
  \draw[ultra thick,red!50] (t_cur) -- ++(#1,0) node[time] (t_cur) {};
}

\newcommand{\nextwave}[1]{
  \path (0,\value{wavenum}) node[left] {#1} node[time] (t_cur) {};
  \addtocounter{wavenum}{-1}
}

\newenvironment{wave}[3][clk]{
  \begin{tikzpicture}[draw=black, yscale=.7,xscale=0.8]
    \tikzstyle{time}=[coordinate]
    \setlength{\unitlength}{1cm}
    \def\wavewidth{#3}
    \setcounter{wavenum}{0}
    \nextwave{#1}
    \foreach \t in {0,1,...,\wavewidth}{
      \draw[dotted] (t_cur) +(0,.5) node[above] {\t} -- ++(0,.4-#2);
      \clki
    }
}{\end{tikzpicture}}

\begin{figure}[t]
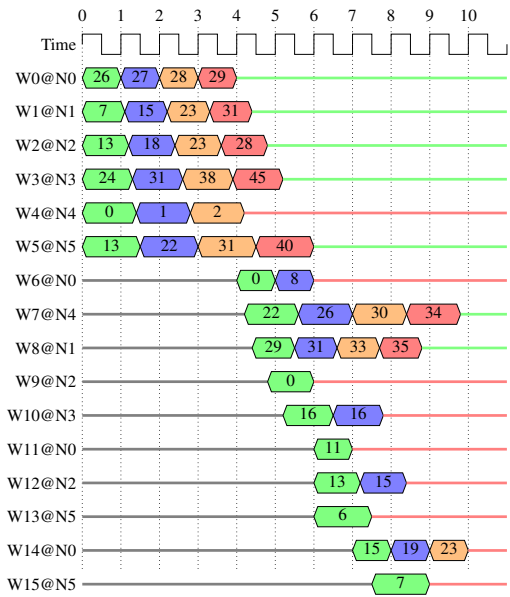

\centering
     \resizebox{0.83\linewidth}{!}{
\begin{wave}[Time]{16}{10}
   \nextwave{W0@N0} \firstphase{26}{1.00} \secondphase{27}{1.00} \thirdphase{28}{1.00} \fourthphase{29}{1.00} \completed{7.00}
   \nextwave{W1@N1} \firstphase{7}{1.10} \secondphase{15}{1.10} \thirdphase{23}{1.10} \fourthphase{31}{1.10} \completed{6.60}
   \nextwave{W2@N2} \firstphase{13}{1.20} \secondphase{18}{1.20} \thirdphase{23}{1.20} \fourthphase{28}{1.20} \completed{6.20}
   \nextwave{W3@N3} \firstphase{24}{1.30} \secondphase{31}{1.30} \thirdphase{38}{1.30} \fourthphase{45}{1.30} \completed{5.80}
   \nextwave{W4@N4} \firstphase{0}{1.40} \secondphase{1}{1.40} \thirdphase{2}{1.40} \killed{6.80}
   \nextwave{W5@N5} \firstphase{13}{1.50} \secondphase{22}{1.50} \thirdphase{31}{1.50} \fourthphase{40}{1.50} \completed{5.00}
   \nextwave{W6@N0} \pending{4.00} \firstphase{0}{1.00} \secondphase{8}{1.00} \killed{5.00}
   \nextwave{W7@N4} \pending{4.20} \firstphase{22}{1.40} \secondphase{26}{1.40} \thirdphase{30}{1.40} \fourthphase{34}{1.40} \completed{1.20}
   \nextwave{W8@N1} \pending{4.40} \firstphase{29}{1.10} \secondphase{31}{1.10} \thirdphase{33}{1.10} \fourthphase{35}{1.10} \completed{2.20}
   \nextwave{W9@N2} \pending{4.80} \firstphase{0}{1.20} \killed{5.00}
   \nextwave{W10@N3} \pending{5.20} \firstphase{16}{1.30} \secondphase{16}{1.30} \killed{3.20}
   \nextwave{W11@N0} \pending{6.00} \firstphase{11}{1.00} \killed{4.00}
   \nextwave{W12@N2} \pending{6.00} \firstphase{13}{1.20} \secondphase{15}{1.20} \killed{2.60}
   \nextwave{W13@N5} \pending{6.00} \firstphase{6}{1.50} \killed{3.50}
   \nextwave{W14@N0} \pending{7.00} \firstphase{15}{1.00} \secondphase{19}{1.00} \thirdphase{23}{1.00} \killed{1.00}
   \nextwave{W15@N5} \pending{7.50} \firstphase{7}{1.50} \killed{2.00}
\end{wave}}
\caption{Metaoptimization with HyperTrick ($r=25\%$, $W_0=16$) on a toy problem, $N_p=4$ phases (in green, blue, orange, and red). The workers $\{Wi\}_{i=0..15}$ are initially scheduled on 6 nodes $\{Ni\}_{i=0..5}$.
The cells show the metric $f(p)$ for each worker and phase $p$; here, $f(p)=ap+b$, where $a$, $b$ are random values.
Green lines indicate completion of all phases, red lines indicate early termination.
The phase execution time is variable: this is common in a heterogeneous systems, or when the hyperparameters affect the computational cost of the underneath optimization problem.
The process takes 10 units of time.}
\label{fig:waveWithHyperTrick}
\end{figure}

\begin{figure}[t]
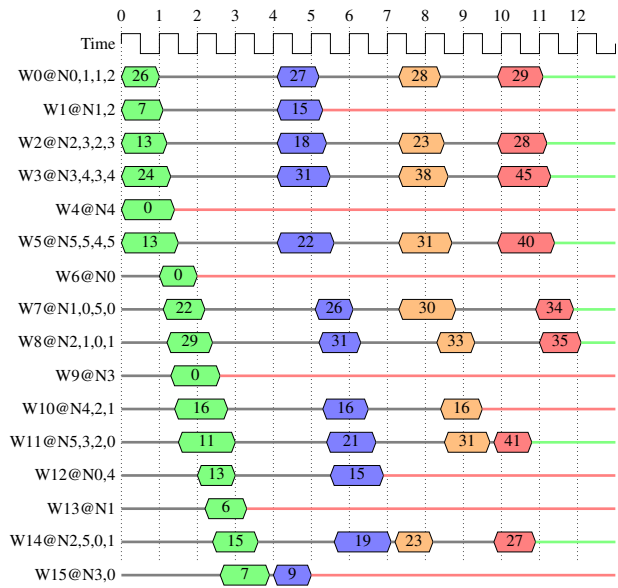

    \centering
     \resizebox{\linewidth}{!}{
\begin{wave}[Time]{16}{12}
   \nextwave{W0@N0,1,1,2} \firstphase{26}{1.00} \pending{3.10} \secondphase{27}{1.10} \pending{2.10} \thirdphase{28}{1.10} \pending{1.50} \fourthphase{29}{1.20} \completed{1.90}
   \nextwave{W1@N1,2} \firstphase{7}{1.10} \pending{3.00} \secondphase{15}{1.20} \killed{7.70}
   \nextwave{W2@N2,3,2,3} \firstphase{13}{1.20} \pending{2.90} \secondphase{18}{1.30} \pending{1.90} \thirdphase{23}{1.20} \pending{1.40} \fourthphase{28}{1.30} \completed{1.80}
   \nextwave{W3@N3,4,3,4} \firstphase{24}{1.30} \pending{2.80} \secondphase{31}{1.40} \pending{1.80} \thirdphase{38}{1.30} \pending{1.30} \fourthphase{45}{1.40} \completed{1.70}
   \nextwave{W4@N4} \firstphase{0}{1.40} \killed{11.60}
   \nextwave{W5@N5,5,4,5} \firstphase{13}{1.50} \pending{2.60} \secondphase{22}{1.50} \pending{1.70} \thirdphase{31}{1.40} \pending{1.20} \fourthphase{40}{1.50} \completed{1.60}
   \nextwave{W6@N0} \pending{1.00} \firstphase{0}{1.00} \killed{11.00}
   \nextwave{W7@N1,0,5,0} \pending{1.10} \firstphase{22}{1.10} \pending{2.90} \secondphase{26}{1.00} \pending{1.20} \thirdphase{30}{1.50} \pending{2.10} \fourthphase{34}{1.00} \completed{1.10}
   \nextwave{W8@N2,1,0,1} \pending{1.20} \firstphase{29}{1.20} \pending{2.80} \secondphase{31}{1.10} \pending{2.00} \thirdphase{33}{1.00} \pending{1.70} \fourthphase{35}{1.10} \completed{0.90}
   \nextwave{W9@N3} \pending{1.30} \firstphase{0}{1.30} \killed{10.40}
   \nextwave{W10@N4,2,1} \pending{1.40} \firstphase{16}{1.40} \pending{2.50} \secondphase{16}{1.20} \pending{1.90} \thirdphase{16}{1.10} \killed{3.50}
   \nextwave{W11@N5,3,2,0} \pending{1.50} \firstphase{11}{1.50} \pending{2.40} \secondphase{21}{1.30} \pending{1.80} \thirdphase{31}{1.20} \pending{0.10} \fourthphase{41}{1.00} \completed{2.20}
   \nextwave{W12@N0,4} \pending{2.00} \firstphase{13}{1.00} \pending{2.50} \secondphase{15}{1.40} \killed{6.10}
   \nextwave{W13@N1} \pending{2.20} \firstphase{6}{1.10} \killed{9.70}
   \nextwave{W14@N2,5,0,1} \pending{2.40} \firstphase{15}{1.20} \pending{2.00} \secondphase{19}{1.50} \pending{0.10} \thirdphase{23}{1.00} \pending{1.60} \fourthphase{27}{1.10} \completed{2.10}
   \nextwave{W15@N3,0} \pending{2.60} \firstphase{7}{1.30} \pending{0.10} \secondphase{9}{1.00} \killed{8.00}
\end{wave}}
\caption{Metaoptimization with a variant of Successive Halving which terminates 25\% of workers at the end of every phase
, for the same toy problem in Fig. \ref{fig:waveWithHyperTrick}. Workers
are dispatched to different nodes over the course of the entire process.
For example $W1@N1,2$ indicates that $W1$ executes the first two phases on nodes 1 and 2, respectively. The process takes 12.1 units of time.}
\label{fig:waveWithSuccessiveHalving}
\end{figure}

For higher clarity, Fig. \ref{fig:waveWithHyperTrick} shows HyperTrick with $W_0 = 16$ workers, $N_p=4$ phases, 6 compute nodes, and a target eviction rate $r=25\%$, on a toy problem.
Accordingly to Eq.~(\ref{eq:n_DCM}), the minimum number of workers allowed to continue at the end of the first, second and third phase, are $W_1^{DCM}=8$, $W_2^{DCM}=6$, and $W_3^{DCM}=4$, respectively.
Initially, 6 workers $\{Wi\}_{i=0..5}$ run 6 optimization experiments on the 6 available nodes, $\{Ni\}_{i=0..5}$.
The fast worker $W0$ terminates all 4 phases at time $t=4$: the node N0 is released and $W6$ starts a new experiment on the same node.
At time $t = 4.2$, $W4$ is the fifth worker (after $W0, ..., W3$) to reach the end of the third phase.
HyperTrick consequently switches from DCM to WSM for the third phase: from now on, each worker completing this phase will continue only if its score is in the top $\sqrt{r}=50\%$ of the scores at the end of the third phase.
The workers $\{Wi\}_{i=0..3}$ won't be affected by this rule as they already started the fourth phase.
Since $W4$ reports a low metric at the end of the third phase, it is terminated by HyperTrick and $N4$ is reallocated to start $W7$.
At time $t=4.5$, $W5$ reaches the end of its third phase; its metric ($31$) is in the top half, thus $W5$ is allowed to proceed to the last phase.
At $t = 6$, 6 workers have already reported their metrics for the first phase, $W6$ completes it and reports a low metric ($8$), thus it is terminated; $N0$ is reallocated for $W11$.
Overall, 10 units of time are required to complete the entire metaoptimization process.

Some of the advantages offered by HyperTrick are evident after analyzing Fig. \ref{fig:waveWithSuccessiveHalving}, that  shows Successive Halving, terminating $25\%$ of the workers at the end of each phase, on the same toy problem. 
When compared to Hypertrick, Successive Halving takes a longer amount of time (12.1 units) and achieves a lower occupancy of the system, because of the need to synchronize workers.
Successive Halving does not allow any slow learner to run to completion (\emph{e.g.}, $W1$ is terminated early, whereas in HyperTrick it runs to completion and achieves a final, above average metric of 31).
In this example, Successive Halving requires workers to support preemption, as experiments are stopped and restarted after a while, potentially on a different node; the overhead for context switches is optimistically assumed to be zero here.
A simplified implementation of Successive Halving that allocates statically each worker to one node is feasible, but it takes more time (15.3 units of time - see Fig. \ref{fig:waveWithSuccessiveHalvingFixedNodeAllocation}, reported in the Appendix for sake of space).

\section{GA3C}
\label{sec:GA3C}

In this Section we introduce some basic concepts of RL and summarize the main characteristics of GA3C~\cite{Bab16,Bab17}, to help the reader interpret the experimental results presented in the next section.

\subsection{Reinforcement Learning and REINFORCE}
In RL, an agent observes a state $s_t$ at time $t$ and selects an action $a_t$, following a policy $\pi$, that maps $s_t$ to $a_t$.
The agent receives then a feedback from the environment in the form of a reward $r_t$.
The goal of RL is to find a policy $\pi$ that maximizes the sum of the expected rewards.

In policy-based, model-free methods, a DNN can be used to compute $\pi(a_t|s_t;\theta)$, where $\theta$ is the set of DNN weights.
Algorithms from the REINFORCE~\cite{williams1992simple} family use gradient ascent on $E[R_t]$, where $R_t=\sum_{i=0}^{\infty}\gamma^ir_{t+i}$ is the  accumulated reward from time $t$, discounted by the factor $\gamma\in (0, 1]$.
Small values of $\gamma$ generate short-sighted agents that prefer immediate rewards, whereas large $\gamma$ values create agents with a long term strategy, but more difficult to train.
The vanilla REINFORCE updates $\theta$ uses the gradient $\nabla_\theta\log\pi(a_t|s_t;\theta)R_t$, an unbiased estimator of $\nabla_\theta E[R_t]$; its variance is reduced by subtracting a learned \textit{baseline} (a function of the state $b_t(s_t)$) and using the gradient $\nabla_\theta\log\pi(a_t|s_t;\theta)[R_t~-~b_t(s_t)]$ instead.
One common baseline is the value function $V^\pi(s_t) = E[{R_t|s_t}]$, which is the expected return for the policy $\pi$ starting from $s_t$.
The policy $\pi$ and the baseline $b_t$ can be viewed as \textit{actor} and \textit{critic} in an actor-critic architecture \cite{Sutton:1998:IRL:551283}.

\subsection{A3C and GA3C}

A3C~\cite{Mni16}, a successful RL actor-critic algorithm, uses a DNN to compute both the policy and value function.
The DNN trained to play Atari games in~\cite{Mni16} has two convolutional layers, one fully connected layer, and ReLU activations.
The DNN outputs are: a softmax layer for the policy function approximation $\pi\left(a_t | s_t; \theta \right)$, and a linear layer for $V \left(s_t; \theta\right)$.
Multiple agents play concurrently and optimize the DNN through asynchronous gradient descent, with the DNN weights stored in a central parameter server.
After each update, the central server propagates new weights to the agents. 
The variance of the critic $V\left(s_t; \theta\right)$ is reduced (at the price of an increased bias) by N-step bootstrapping:
the agents send updates to the server after every $t_{max}$ actions, or when a terminal state is reached.
The cost function for the policy is:
\begin{equation}
 \log\pi\left(a_t | s_t; \theta \right)
\left[\tilde{R}_t - V\left(s_t; \theta_t \right) \right]
+ \beta H \left[\pi \left(s_t; \theta \right)\right],
\label{eq:costPi}
\end{equation}
where $\theta_t$ are the DNN weights $\theta$ at time $t$, $\tilde{R}_t = \sum_{i=0}^{k-1}{\gamma^i r_{t+i} + \gamma^k V\left(s_{t+k}; \theta_t \right)}$ is the bootstrapped discounted reward in the time interval from $t$ to $t+k$ and $k$ is upper-bounded by $t_{max}$, and $H\left[\pi \left(s_t; \theta \right)\right]$ is an entropy term to favor exploration, whose importance is modulated by $\beta$.
In the original A3C paper, $t_{max}$ is empirically set to 5, which experimentally achieves convergence for most of the Atari games in a reasonable amount of time; nonetheless, the optimal bias-variance trade-off for the critic may be achieved for other values of $t_{max}$, depending on the specific game.
The cost function for the estimated value function is:
\begin{equation}
\left[\tilde{R}_t - V \left( s_t; \theta \right) \right]^2,
\label{eq:costV}
\end{equation}
which uses again the bootstrapped estimate $\tilde{R}_t$. The server collects gradients $\nabla \theta$ from both of the cost functions and uses the standard non-centered RMSProp~\cite{tieleman2012lecture} to optimize them.
The gradients of the two cost functions can be either shared or separated  between agent threads, but the shared implementation is known to be more robust \cite{Mni16}.
%
%

A3C~\cite{Mni16} uses $16$ agents on a $16$ core CPU and takes four days to learn an Atari game~\cite{OpenAIGym}.
Its GPU version, GA3C, is only slightly different from A3C (see details in \cite{Bab16,Bab17}), but it reaches convergence in about one fourth of the time and it is therefore the one adopted here for our experiments.
\section{Results and Discussion}
\label{sec:ResultsAndDiscussion}

\subsection{Experimental setup}
\label{subsec:experimental_setup}

We implement HyperTrick (with settings reported in Table \ref{table:hypertricksettings}) on Maglev to learn to play four different Atari games (Boxing, Pong, Ms-Pacman, and Centipede) in the OpenAI Gym environment~\cite{Bro16}, through GA3C. Our system has 300 compute nodes, each including a dual 20-core Intel Xeon 
E5-2698 v4 2.2 GHz CPU and 8 Nvidia V100 GPUs. 
HyperTrick optimizes three hyperparameters: learning rate, $\gamma$, and $t_{max}$.
A large learning rate can lead to training instabilities, whereas for a small one the convergence is slow and the capability to evade from local minima is limited. 
The discount factor $\gamma$ determines the short/far-sightedness of the agent; a small $\gamma$ leads to agents that easily learn a sub-optimal policy which maximizes immediate returns, but lack from a long term strategy; a large $\gamma$ generates agents that weigh future rewards more, but training is more difficult.
The hyperparameter $t_{max}$ affects both the convergence properties and the computational cost of GA3C: a large $t_{max}$ leads to high variance estimates of $\tilde{R}_t$ and consequently high variance updates of the cost functions in Eqs. (\ref{eq:costPi}) and (\ref{eq:costV}); increasing $t_{max}$ also increases the batch size, which leads to a better utilization of the GPU in GA3C, but decreases the number of policy updates per second, since a large number of frames have to be played to populate the batch.
Decreasing $t_{max}$ reduces the variance of $\tilde{R}_t$, but increases the bias; it also reduces the batch size, eventually leading to a higher number of biased updates per second.

In our experiments we run HyperTrick on a population of 100 workers, whose hyperparameters are randomly picked.
The learning rate is sampled from a random log uniform distribution over the [1e-5, 1e-2] interval; $t_{max}$ is sampled from a random quantized log uniform distribution over the [2, 100] interval, with an increment of 1 to pick integer values;
$\gamma$ is sampled uniformly from the $\{0.9, 0.95, 0.99, 0.995, 0.999, 0.9995, 0.9999\}$ set.

\begin{table*}[t]
\centering
  \begin{tabular}{|c|c|c|c|c||c|c|}
    \hline
    Game & Episodes per Phase & $N_p$ & $r$ & $\alpha$ (min[$\alpha$], E[$\alpha$]) & Score (GA3C) & Score (HyperTrick) \\ \hline
    Boxing & 2500 & 10 & 25\% &
    48.2\% (18.87\%, 37.75\%) & 92 & 98\\
    Centipede & 2500 & 10 & 25\% &
    52.2\%
    (18.87\%, 37.75\%) & 7386 & 8707\\
    Ms Pacman & 2500 & 10 & 25\% &
    46.1\% (18.87\%, 37.75\%) & 1978 & 2112\\
    Pong & 2500 & 5 & 25\% &
    59.1\% (30.51\%, 61.02\%) & 18 & 18\\
    \hline
  \end{tabular}
\caption{HyperTrick parameters (episodes per phase; number of phases, $N_p$; target eviction rate, $r$) for metaoptimization on four Atari games. The fifth column indicates the measured ($\alpha$), minimum (min[$\alpha$]) and expected (E[$\alpha$]) worker completion rate achieved by HyperTrick; $100\%$ corresponds to running all the workers to completion.
The last two columns report the score achieved with a standard implementation of GA3C~\cite{Bab17} and that obtained with HyperTrick.
}
\label{table:hypertricksettings}
\end{table*}

\subsection{Metaoptimization}

\subsubsection{Metaoptimization Results}
Table \ref{table:hypertricksettings} reports the scores achieved by HyperTrick, and compares it with the score in~\cite{Bab17} for GA3C and an ``optimal'' hyperparameter configuration, identified by a trial-and-error, for the same Atari games.
HyperTrick consistently achieves comparable scores, demonstrating its effectiveness to identify an optimal policy, without any significant user intervention to set the hyperparameters.

\subsubsection{Worker Selection Analysis}
\begin{figure*}[t]
    \centering
    \subfloat{\includegraphics[width=0.2\textwidth]{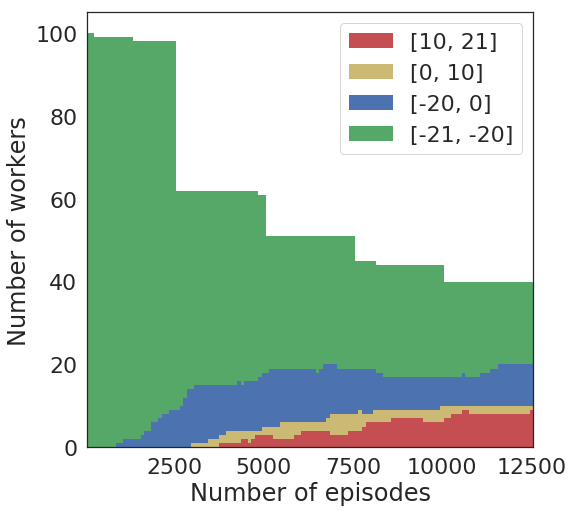}}
    \subfloat{\includegraphics[width=0.2\textwidth]{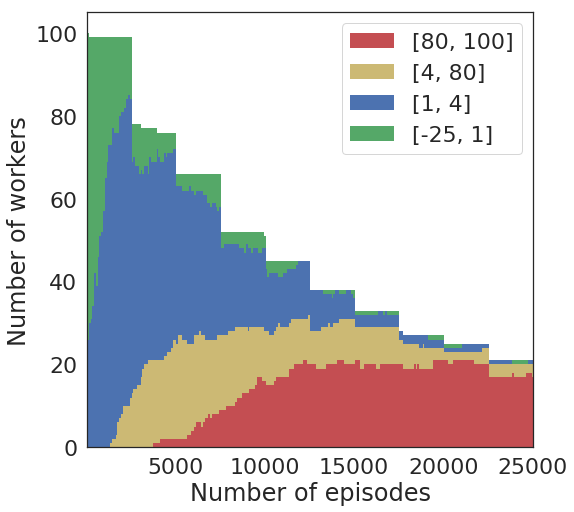}}
    \subfloat{\includegraphics[width=0.2\textwidth]{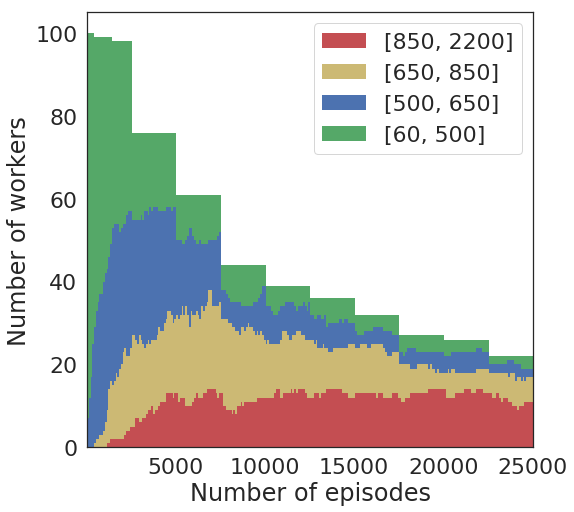}}
    \subfloat{\includegraphics[width=0.2\textwidth]{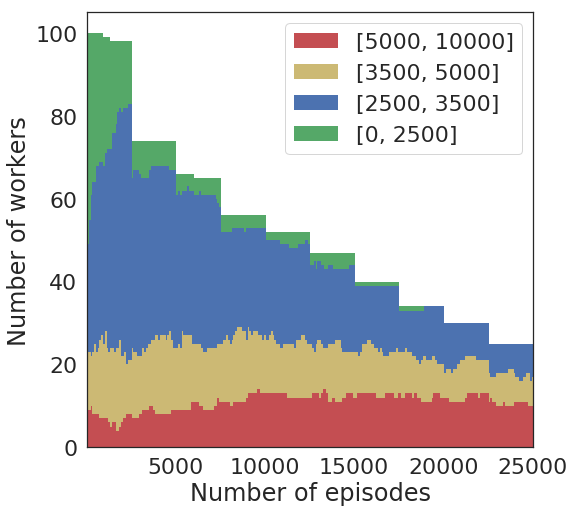}}\\
    \setcounter{subfigure}{0}
    \subfloat[Pong]{\includegraphics[width=0.2\textwidth]{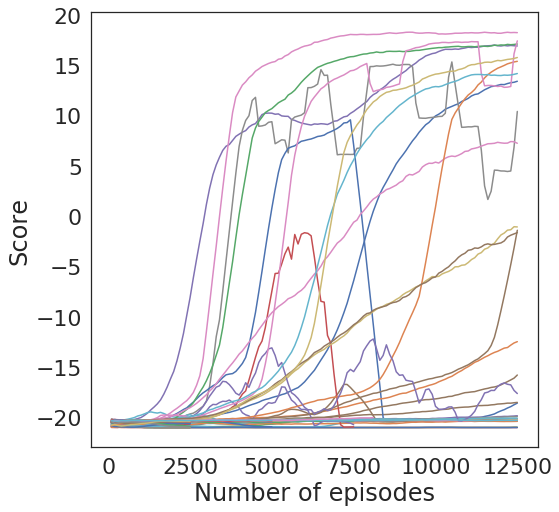}}
    \subfloat[Boxing]{\includegraphics[width=0.2\textwidth]{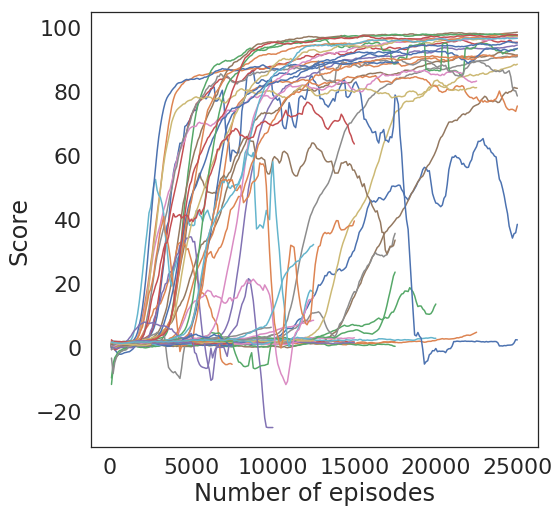}}
    \subfloat[Ms-Pacman]{\includegraphics[width=0.2\textwidth]{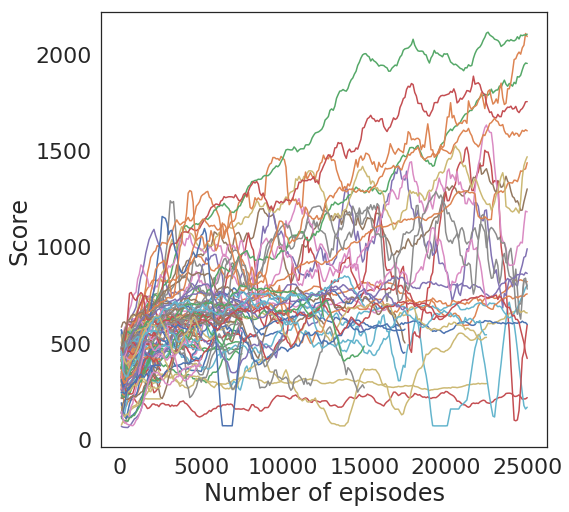}}
    \subfloat[Centipede]{\includegraphics[width=0.2\textwidth]{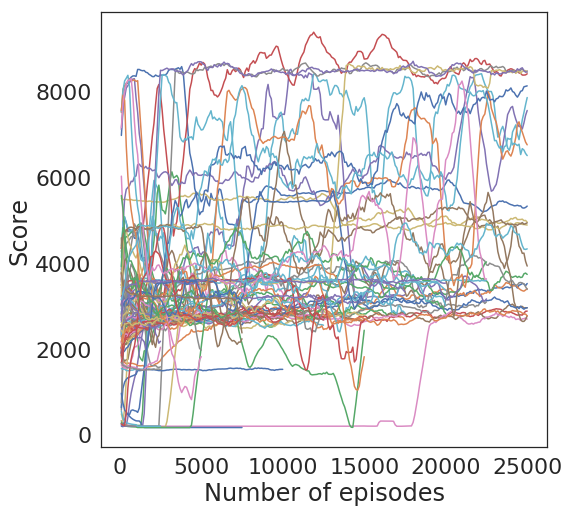}}
    \caption{The first row shows the distribution of the score during metaoptimization with HyperTrick, for four different Atari games learned through GA3C. The lower row shows the learning curves for the entire population of workers.}
    \label{fig:scoreVsSteps}
\end{figure*}

The first row of Fig. \ref{fig:scoreVsSteps} shows the distribution of the metrics (Atari game scores) during our experiments with HyperTrick.
The number of active workers drops after each phase.
Few workers stop before reaching the end of a phase (small drops in Fig. \ref{fig:scoreVsSteps}): they crash or hang for different reasons, but do not affect the output of HyperTrick.
As expected, workers with a low metric are gradually eliminated and the fraction of those achieving a high score increases during  metaoptimization.
More in detail, computationally efficient workers (\emph{e.g.} those with a $t_{max}$ that maximizes the frame generation rate) generally continue to the next phase, independently from their score, because they reach the end of each phase early, with HyperTrick in DCM. 
Computationally demanding workers arrive late at the end of a phase, and they continue only if their score is sufficiently high - in other words, only if they are sample efficient.
Fig.~\ref{fig:scoreVsSteps} also shows the learning curves for the entire set of workers; we notice that unstable training processes (that are common for sub-optimal choices of the hyperparameters, \emph{e.g.} a large learning rate) have a high chance of being eliminated, possibly because of the noise in the reported score at the end of each phase.
In practice, computationally efficient workers are allowed to explore a given hyperparameters configuration in depth, while workers with a higher computational cost must be sample efficient and stable to pass to the next phase.
Workers that are computationally and sample inefficient at the same time are terminated soon.

\begin{figure*}[h!]
    \captionsetup[subfigure]{labelformat=empty}
    \centering
    \subfloat[----- Pong -----]{\includegraphics[width=0.75\textwidth,clip=true,trim=0cm 0cm 0cm 0cm]{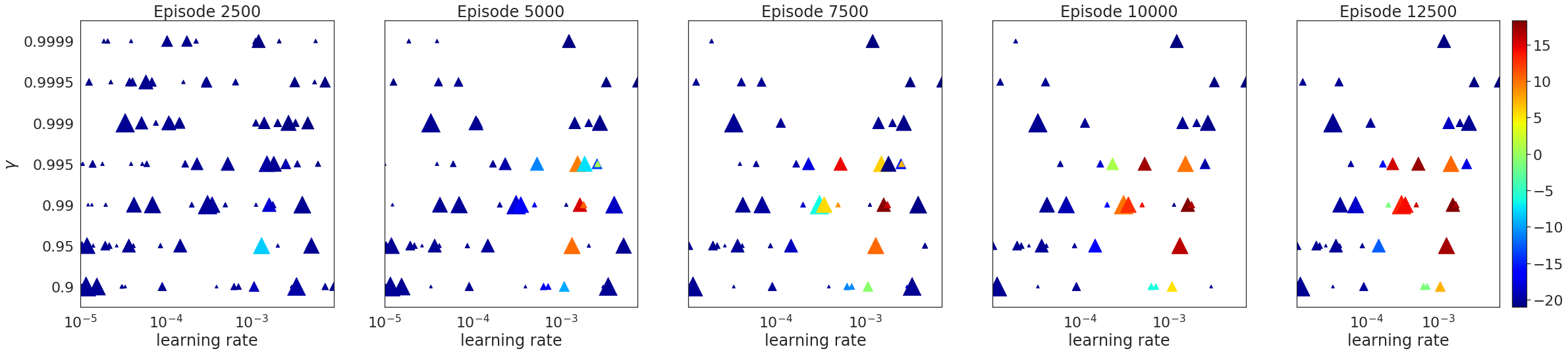}}\\
    \vspace{-0.3cm}
    \subfloat[----- Boxing -----]{\includegraphics[width=0.75\textwidth,clip=true,trim=0cm 0cm 0cm 0cm]{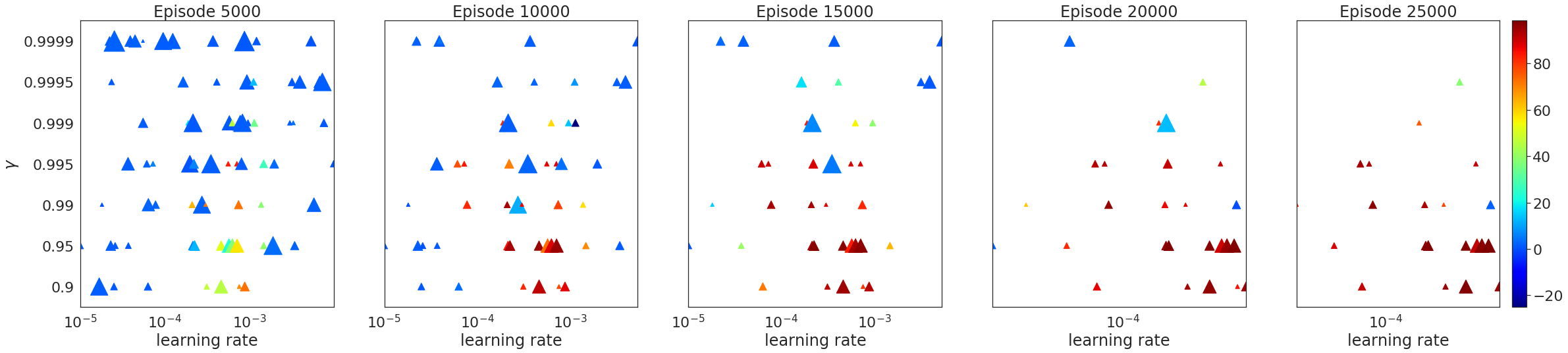}}\\
    \vspace{-0.3cm}
    \subfloat[----- Ms-Pacman ----- ]{\includegraphics[width=0.75\textwidth,clip=true,trim=0cm 0cm 0cm 0cm]{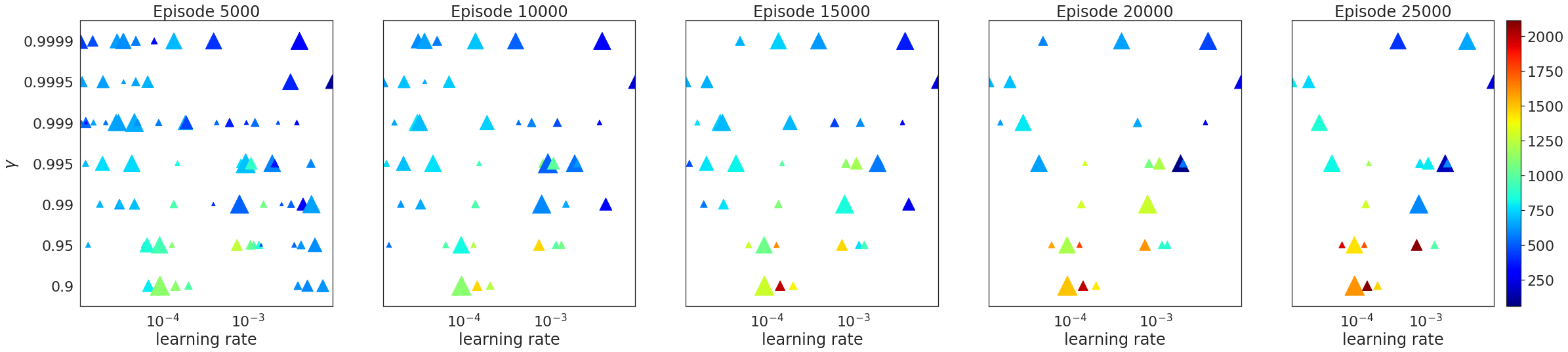}}\\
    \vspace{-0.3cm}
    \subfloat[----- Centipede -----]{\includegraphics[width=0.75\textwidth,clip=true,trim=0cm 0cm 0cm 0cm]{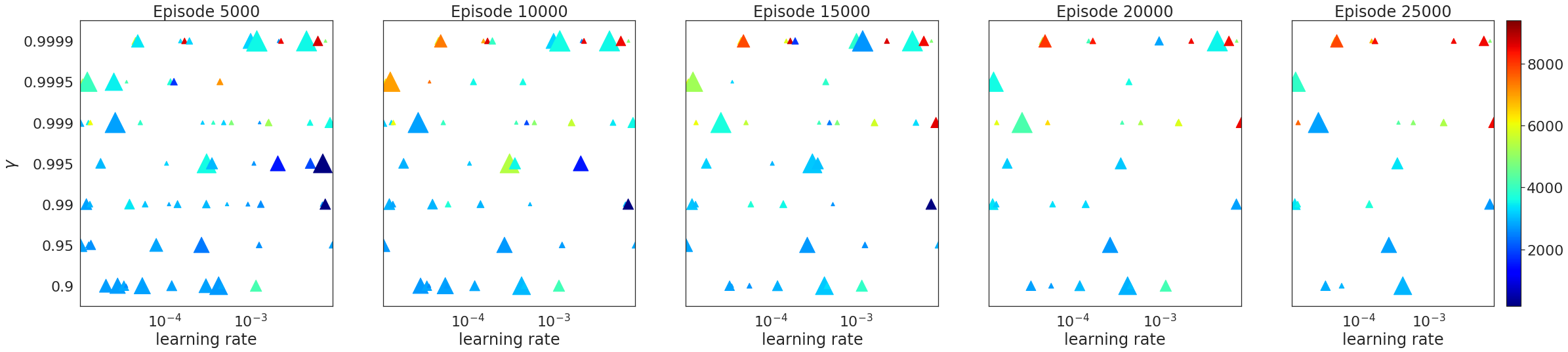}}
    \caption{Selection of the optimal hyperparameters through HyperTrick.
    Each row represents a different Atari game learned with GA3C.
    Each triangle is a worker with a different hyperparameters configuration (learning rate on the x axis, discount $\gamma$ on the y axis, the size of the triangle is proportional to $t_{max}$).
    Workers are terminated by HyperTrick from left to right.}
    \label{fig:hyperparametersSelection}
\end{figure*}

Fig. \ref{fig:hyperparametersSelection} shows the active workers in the hyperparameter space in different phases.
In general, each game has one optimal configuration of learning rate, $\gamma$, and $t_{max}$, which is effectively identified by HyperTrick.
Workers with large $t_{max}$ and low scores are terminated in the first phases of the metaoptimization process: these are the computationally intensive, slow learners, that often reach the end of a phase in WSM mode.

\subsubsection{Worker Completion Rate}

We define the \emph{worker completion rate}, $\alpha$, as the fraction of phases completed by a metaoptimization algorithm.
In the case of Grid Search, with no early stopping, $\alpha=100\%$.
A low value of $\alpha$ indicates that most of the workers have been terminated early;
for the same result obtained on the underneath optimization problem by two metaoptimization algorithms, the one with the lowest $\alpha$ identifies promising hyperparameters configurations more effectively.
Based on the expected (Eq.~(\ref{eq:n_p})) and mininum (Eq.~(\ref{eq:n_DCM})) number of workers in each phase, we compute the minimum and expected $\alpha$ for HyperTrick as:
\begin{eqnarray}
    \textrm{min}[\alpha] =  \frac{\sum\limits_{p=0}^{ N_p-1}{W_p^{DCM}}}{N_pW_0} =
    \frac{\sum\limits_{p=0}^{ N_p-1}{(1-\sqrt{r})(1-r)^p}}{N_p} \nonumber &\\
    = (1-\sqrt{r}) [1-(1-r)^{N_p}]/(rN_p), &\\
    \textrm{E}[\alpha] = \frac{\sum\limits_{p=0}^{ N_p-1}{(1-r)^p}}{N_p} = \frac{1-(1-r)^{N_p}}{rN_p}.
    \label{eq:E_alpha}&
\end{eqnarray}
Table \ref{table:hypertricksettings} shows that, for HyperTrick, $\alpha$ is experimentally close to its expectation for Pong, and slightly higher in other cases, suggesting that the experimental eviction rate is generally lower than the target $r$.
The visual inspection of the learning curves in Fig. \ref{fig:scoreVsSteps} gives a possible interpretation of this phenomenon: these curves are mostly regular and monotonic for Pong, and more irregular for the other games.
In this last case, HyperTrick performs a noisy worker selection in WSM, which may increase the chance of slow or sample inefficient workers to pass the first selections, just to be eliminated in the next phases.
On the other hand, the minimum completion rate $\textrm{min}[\alpha]$ is hardly achievable in practice, as in this case HyperTrick in WSM should terminate all workers.
This represents an upper bound for $\alpha$, which could be approached by using any a-priori knowledge about hyperparameters and their effect on the learning curve.
This may lead to optimal scheduling planning in HyperTrick - since this topic goes beyond the scope of our paper, we leave it for future investigation.
Notice that $\textrm{E}[\alpha]$ for HyperTrick is also the exact completion rate for a vanilla implementation of Successive Halving, assuming no overhead for context switching; HyperTrick  achieves an experimentally higher $\alpha$ in our tests, but it is expected to complete the metaoptimization earlier and to achieve a higher system occupancy (c.f. Figs.~\ref{fig:waveWithHyperTrick} and~\ref{fig:waveWithSuccessiveHalving}), since it does not synchronize the workers.
Experimental evidence of this is provided in Section~\ref{subsubsec:comparison_against_hyperband}.



\subsubsection{Comparison Against Hyperband}
\label{subsubsec:comparison_against_hyperband}

We compare HyperTrick and Hyperband~\cite{Li16}, a recently proposed metaoptimization algorithm which calls Successive Halving multiple times as a sub-routine, in the attempt of automatically finding the optimal balance between breadth and depth of the search.
For Hyperband, we set $\eta=3$ (as in~\cite{Li16}) and $R=27$, which leads to a total of $4$ brackets $s=\{3,2,1,0\}$ and $27 + 9 + 6 + 4 = 46$ configurations of hyperparameters explored by Hyperband (first row in Table ~\ref{table:hyperbandsettings}), initialized randomly.
We define a unit of computational resource ($r_{i,s}$ in Table~\ref{table:hyperbandsettings}) as a set of 500 training episodes, such that the maximum number of training episodes in one phase of Successive Halving is equal to $13,500$.
Since each bracket in Hyperband represents an independent instance of Successive Halving, we run the four brackets in parallel on Maglev.
Experiments are restarted from the first iteration in each phase of Successive Halving.
For this configuration of Hyperband, we can compute the worker completion rate for each bracket (reported in Table~\ref{table:hyperbandsettings}) as $\alpha_s = n_{0,s}R / \sum_i (n_{i,s} r_{i,s})$; for the entire Hyperband algorithm we have $\alpha = \sum_s (n_{0,s}R) / \sum_s \sum_i (n_{i,s} r_{i,s}) = 32.61\%$.
We run Hyperband on 46 nodes and guarantee that all workers start at the same time, with no delay.
For a fair comparison, we run HyperTrick on the same 46 configurations of hyperparameters, on the same nodes, and $N_p = 27$ phases; we compute the target eviction rate of HyperTrick to guarantee that the overall compute time is similar for the two metaoptimization algorithms.
Since both the algorithms analyze the same number of hyperparameter configurations, this is achieved by setting the expected worker completion rate of HyperTrick equal to that of Hyperband, i.e. $E[\alpha] = 32.61\%$, and iteratively solving Eq.(\ref{eq:E_alpha}) for $r$, to get $r= 10.82\%$. 


\begin{table}[]
    \centering
    \begin{tabular}{|c|c|c|c|c|}
    \hline
     & $s=3$ & $s=2$ & $s=1$ & $s=0$ \\
    $i$ & $n_{i,3}$ $r_{i,3}$ & $n_{i,2}$ $r_{i,2}$ & $n_{i,1}$ $r_{i,1}$ & $n_{i,0}$ $r_{i,0}$ \\ \hline
    $0$ & $27$ $1$ & $9$ $3$ & $6$ $9$ & $4$ $27$ \\
    $1$ & $9$ $3$ & $3$ $9$ & $2$ $27$ & --- \\
    $2$ & $3$ $9$ & $1$ $27$ & --- & --- \\
    $3$ & $1$ $27$ & --- & --- & --- \\
    \hline
    $\alpha_s$ & $14.81\%$ & $33.33\%$ & $66.67\%$ & $100\%$\\
    \hline
  \end{tabular}
\caption{The Hyperband configuration used in testing ($\eta=3$,~$R=27$) leads to the definition of 4 brackets $s=\{0,1,2,3\}$, each $(n_{i,s}, r_{i,s})$ corresponding to an instance of Successive Halving; $n_{i,s}$ indicates the number of experiments running in the i-th phase of Successive Halving; whereas $r_{i,s}$ indicates the computational resources allocated for each experiment. In the specific case considered here, $r_{i,s} = 1$ corresponds to 500 training episodes while running GA3C. The worker completion rate $\alpha_s$ is also indicated for each individual bracket, whereas it is equal to $32.61\%$ for the entire Hyperband algorithm.}
\label{table:hyperbandsettings}
\end{table}

\begin{figure*}[th!]
    \centering
    \subfloat{\includegraphics[width=0.24\textwidth]{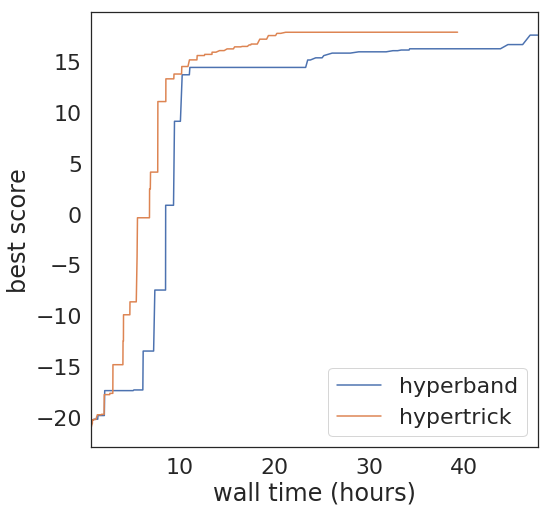}}
    \subfloat{\includegraphics[width=0.24\textwidth]{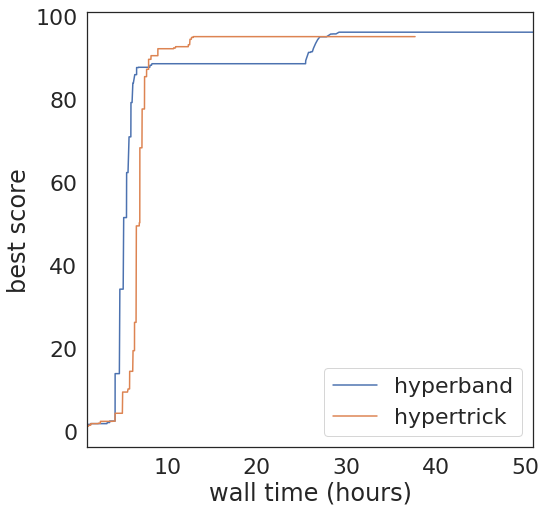}}
    \subfloat{\includegraphics[width=0.24\textwidth]{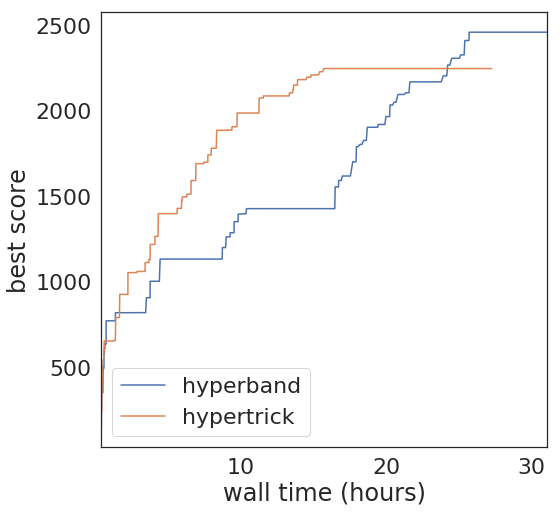}}
    \subfloat{\includegraphics[width=0.24\textwidth]{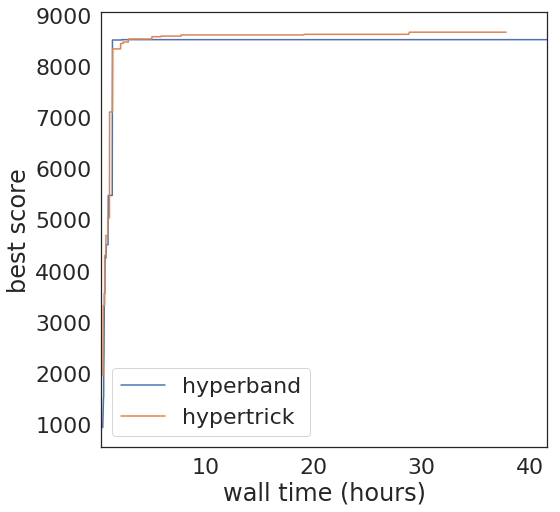}}\\
    \subfloat[Pong]{\includegraphics[width=0.24\textwidth]{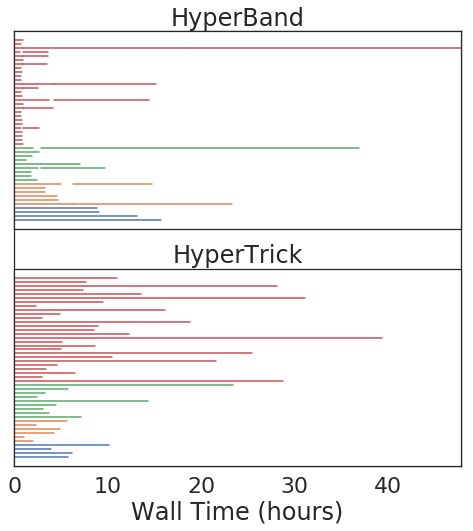}}
    \subfloat[Boxing]{\includegraphics[width=0.24\textwidth]{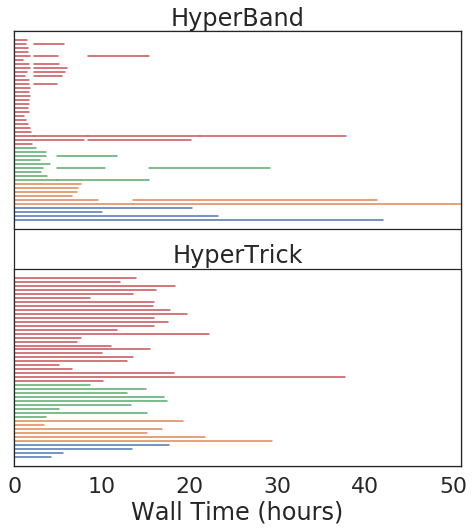}}
    \subfloat[Ms-Pacman]{\includegraphics[width=0.24\textwidth]{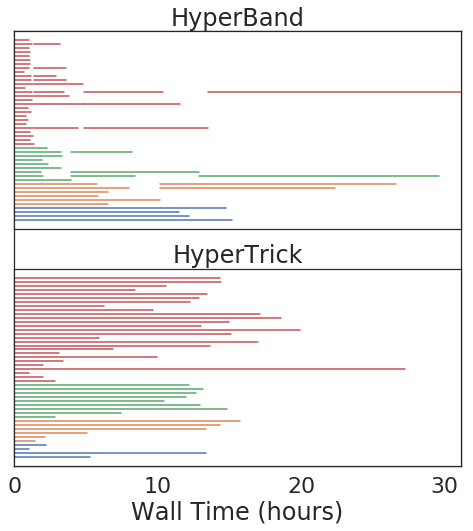}}
    \subfloat[Centipede]{\includegraphics[width=0.24\textwidth]{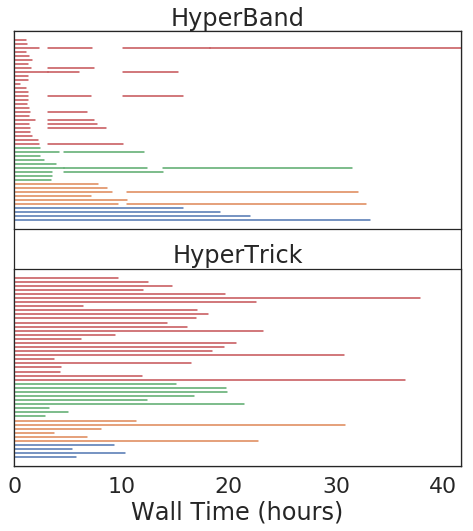}}\\
    \setcounter{subfigure}{0}
    \subfloat{\includegraphics[width=0.24\textwidth]{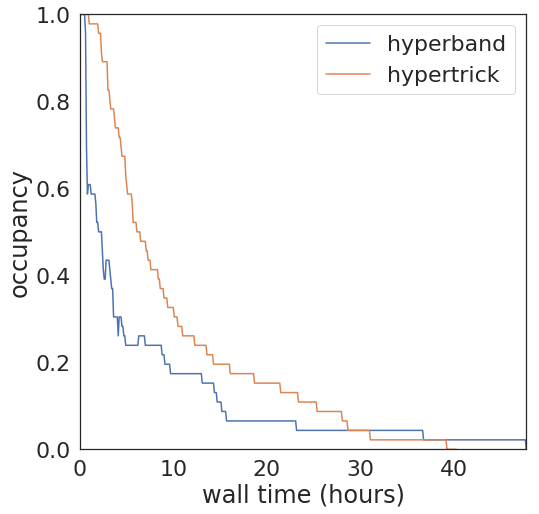}}
    \subfloat{\includegraphics[width=0.24\textwidth]{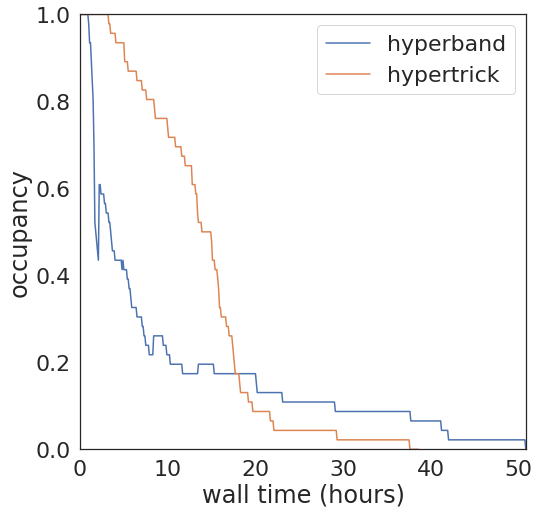}}
    \subfloat{\includegraphics[width=0.24\textwidth]{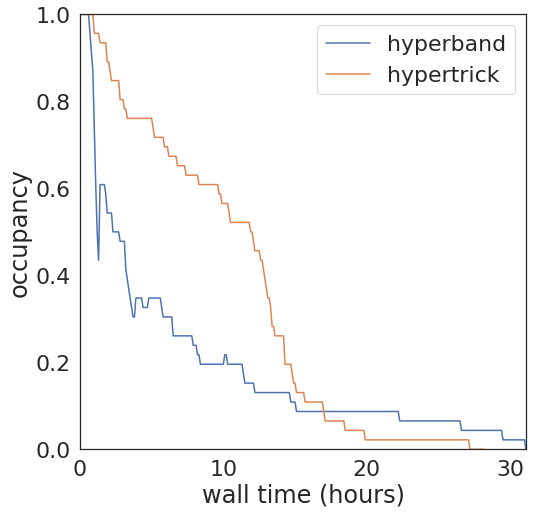}}
    \subfloat{\includegraphics[width=0.24\textwidth]{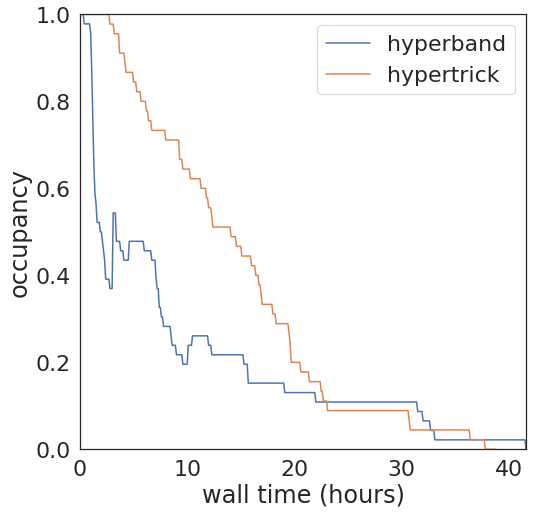}}
    \caption{The first row shows the score of the best agent as a function of the wall time on four different Atari games for Hyperband and HyperTrick; the performances of the two metaoptimization algorithms are comparable in these graphs. The middle row shows the execution timelines for each worker; each color corresponds to a bracket of Hyparband - the same color is used in HyperTrick for an easy comparison. Workers may be idle in the Hyperband case because of the synchronization required at the and of each phase in Successive Halving. The last row shows the occupancy of the nodes of the distributed system as a function of the wall time; the occupancy is generally higher for HyperTrick.}
    \label{fig:hyperbandVsHypertrick}
\end{figure*}

Experimental results are summarized in Table \ref{table:hyperbandVsHyperTrickResults} and illustrated in Fig.~\ref{fig:hyperbandVsHypertrick}.
Hyperband and HyperTrick identify the same optimal configuration of hyperparameters for Pacman and Boxing; the slightly better score reported for Hyperband in Table \ref{table:hyperbandVsHyperTrickResults} is due to the non-deterministic nature of the training and evaluation procedures.
In the case of Pong and Centipede the hyperparameter configurations are different, but the final score similar, possibly because multiple hyperparameter configurations can lead to similar results in this case (see also Fig.~\ref{fig:hyperparametersSelection}). 
Despite the fact that HyperTrick and Hyperband have the same expected $\alpha$, and therefore execute (on average) the same overall number of operations, HyperTrick generally terminates the metaoptimization procedure in a shorter amount of time.
The last row of Fig.~\ref{fig:hyperbandVsHypertrick} highlights that HyperTrick achieves a higher occupancy of the computational nodes in the distributed system, which explains the overall shorter time; this is a direct effect of the lack of synchronization in HyperTrick, which immediately reallocates a node for a new experiment when a worker is terminated, whereas Successive Halving in each bracket of Hyperband pays an additional overhead due to phase synchronization, that leads some of the workers to remain idle (middle row of Fig.~\ref{fig:hyperbandVsHypertrick}).
Table~\ref{table:hyperbandVsHyperTrickResults} also highlights that HyperTrick generally identifies the best configuration in a significantly shorter amount of wall time, compared to Hyperband.
The only exception to this is the case of Centipede in Table~\ref{fig:importanceOfHyperparameters}, but  Fig.~~\ref{fig:hyperbandVsHypertrick} reveals that the best configuration identified by HyperTrick is associated to a late, slight oscillation of the maximum score, while both HyperTrick and Hyperband effectively identify a close-to-optimal solution in a similar amount of time.
Another practical issue (actually observed during our experiments) in HyperBand is that a single point of failure may jeopardize the entire Successive Halving bracket, since workers need to wait for each other.
This is not the case with HyperTrick due to the absence of synchronization points.

\begin{table*}[t]
\centering
  \begin{tabular}{|c|c|c|c|c|ccc|}
    \hline
    Game & Method & Best Score & Total Wall Time & Time To Best Score & & Best Config &\\
    &&&&& $LR$ & $\gamma$ & $T_{max}$\\
    \hline
    Boxing    & HyperBand  & 96   & 51h & 29h & $3.3e^{-4}$ & $0.99$ & $13$\\
              & HyperTrick & 95   & 38h & 13h & $3.3e^{-4}$ & $0.99$ & $13$\\ \hline
    Centipede & HyperBand  & 8521 & 42h & 2h  & $5.4e^{-3}$ & $0.9995$ & $72$ \\
              & HyperTrick & 8667 & 38h & 29h & $1.2e^{-4}$ & $0.9999$ & $33$ \\ \hline
    Pacman    & HyperBand  & 2456 & 31h & 26h & $1.6e^{-4}$ & $0.95$ & $73$ \\
              & HyperTrick & 2243 & 27h & 16h & $1.6e^{-4}$ &  $0.95$ & $73$ \\ \hline
    Pong      & HyperBand  & 17.5 & 48h & 47h & $2.0^{-3}$ & $0.95$ & $64$ \\
              & HyperTrick & 17.8 & 39h & 22h & $5.9e^{-4}$ & $0.995$ & $6$ \\
    \hline
  \end{tabular}
\caption{HyperBand vs HyperTrick results on four Atari games.}
\label{table:hyperbandVsHyperTrickResults}
\end{table*}

\subsection{RL Training Results}

\begin{figure*}[th!]
    \centering
    \subfloat{{\includegraphics[width=0.22\textwidth,clip=true,trim=0cm 0cm 0cm 0cm]{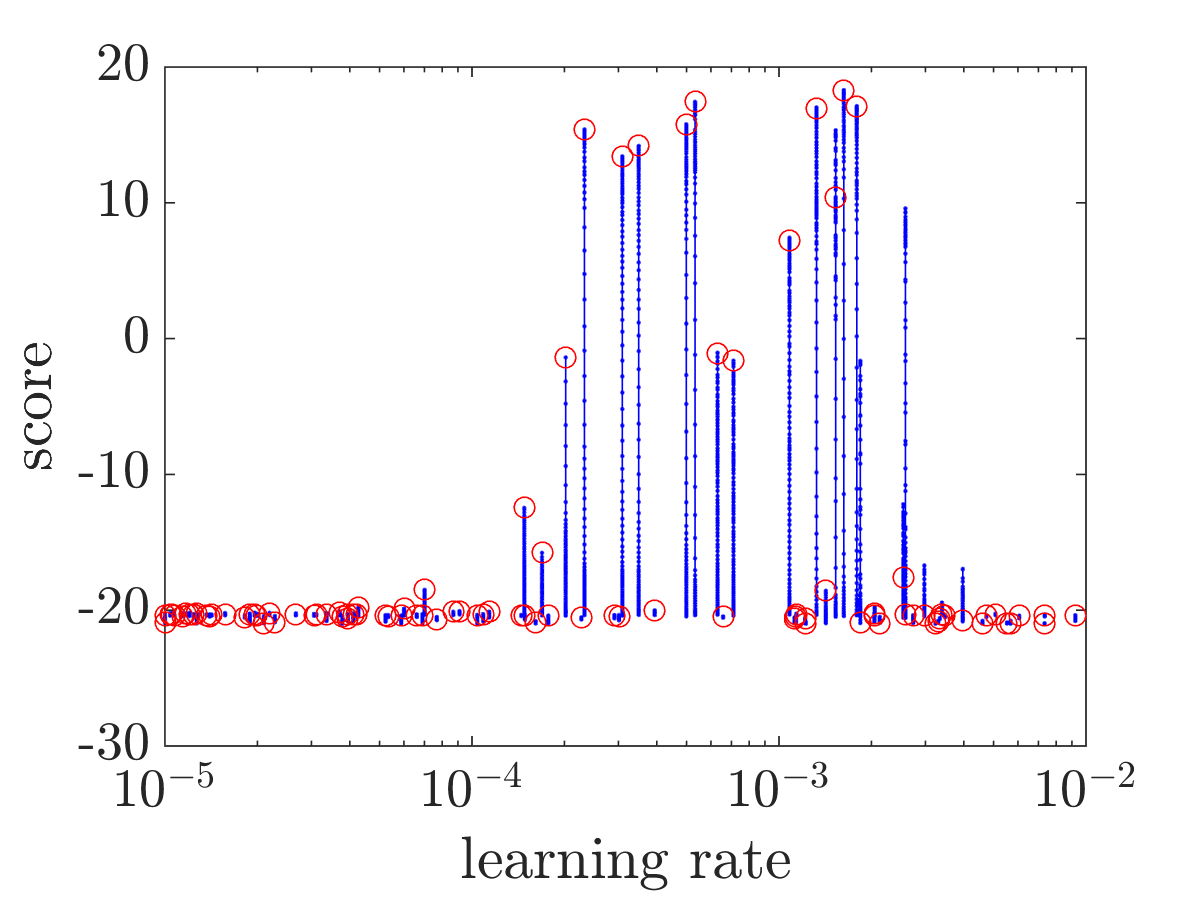}}}
    \subfloat{{\includegraphics[width=0.22\textwidth,clip=true,trim=0cm 0cm 0cm 0cm]{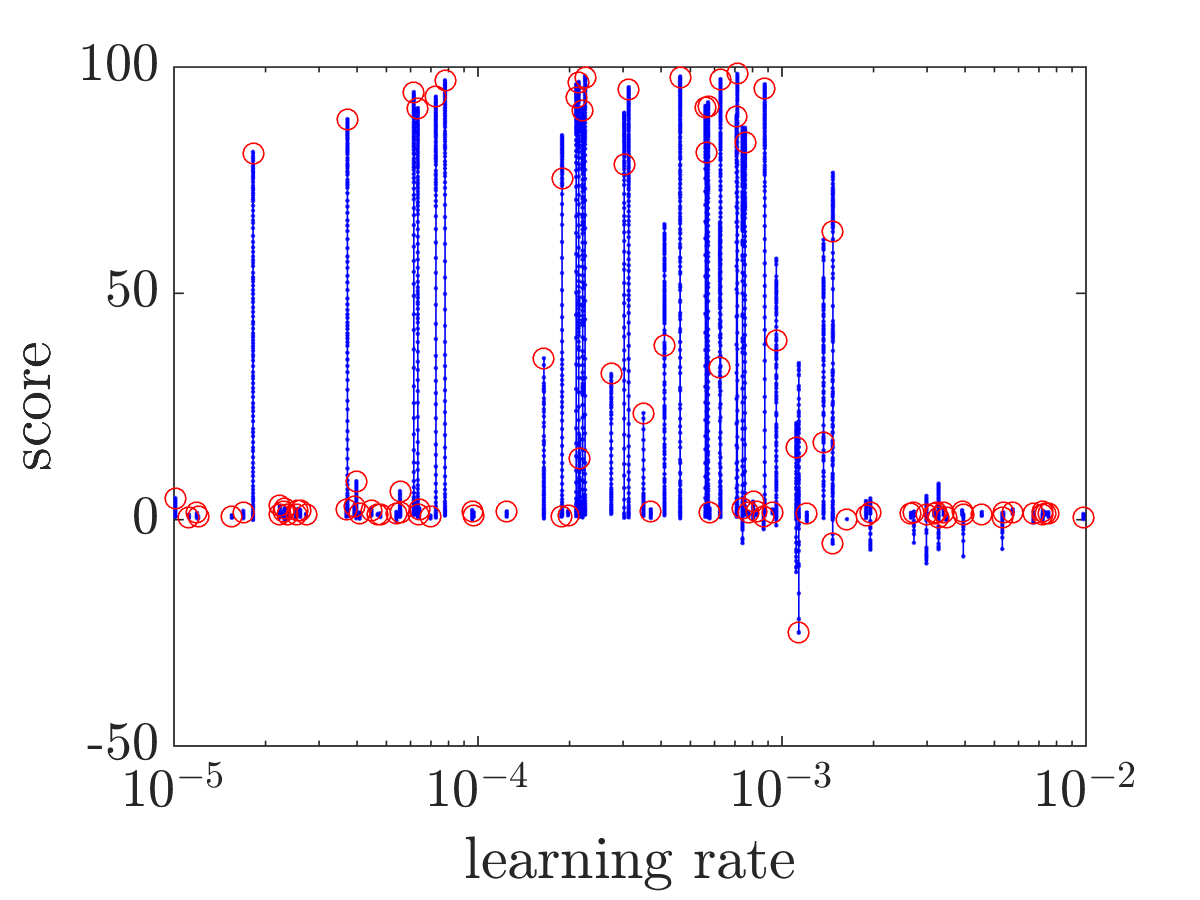}}}
    \subfloat{{\includegraphics[width=0.22\textwidth,clip=true,trim=0cm 0cm 0cm 0cm]{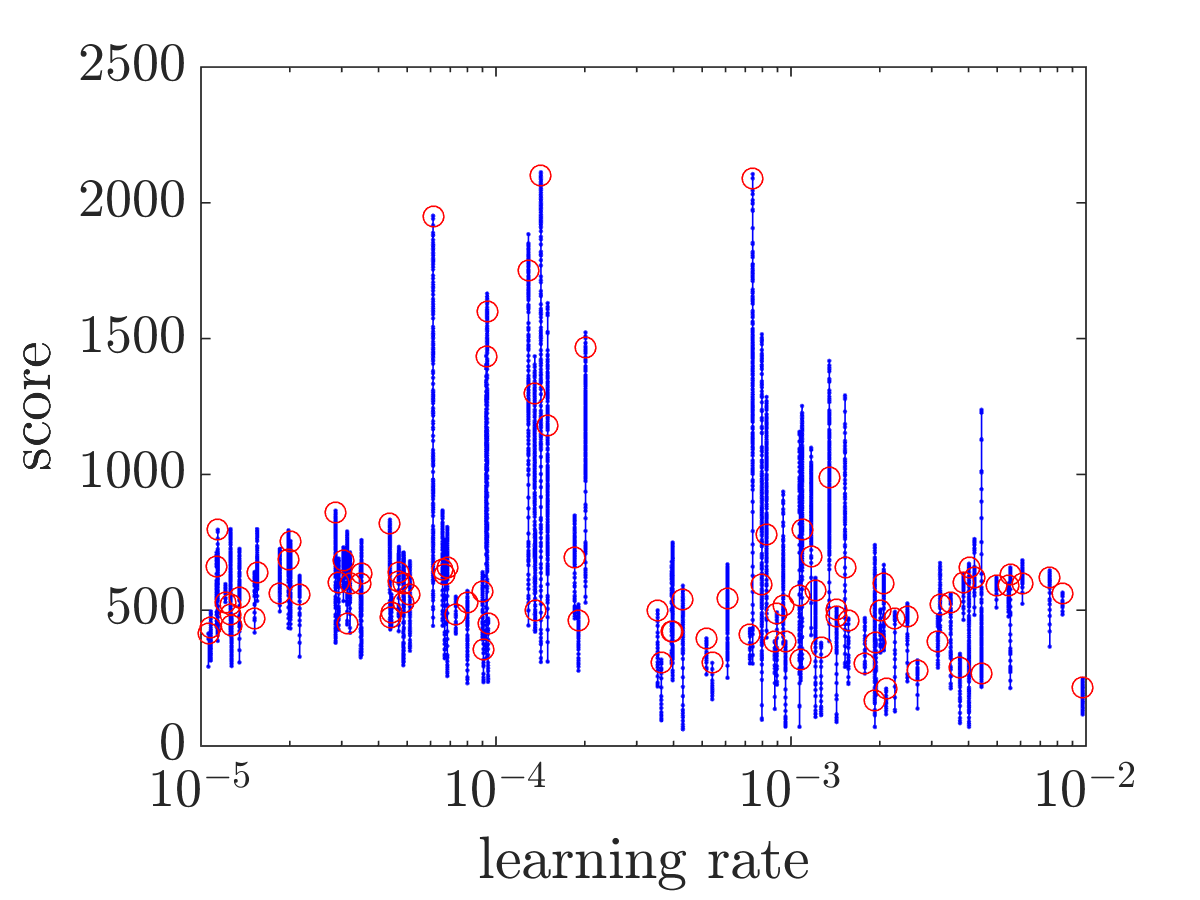}}}
    \subfloat{{\includegraphics[width=0.22\textwidth,clip=true,trim=0cm 0cm 0cm 0cm]{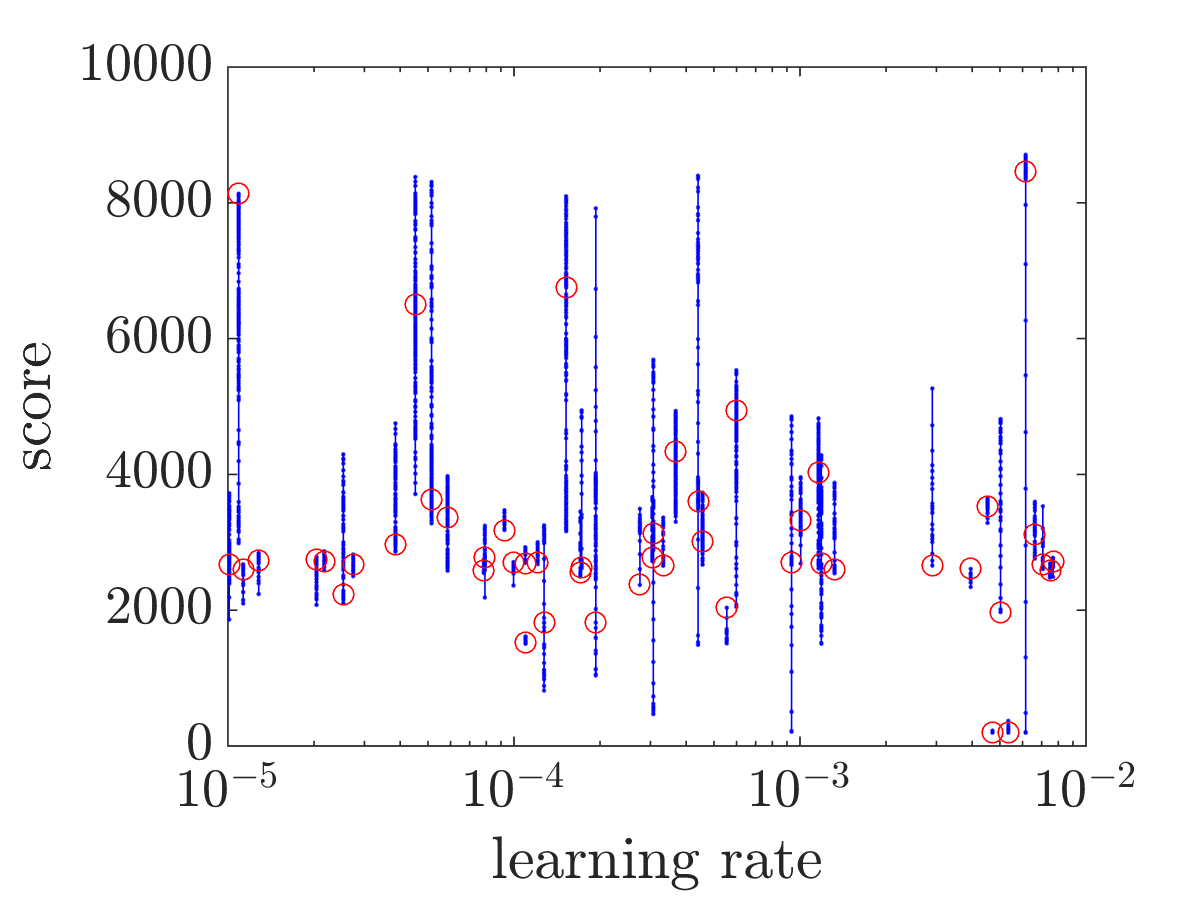}}}\\
    \vspace{-0.4cm}
    \subfloat{{\includegraphics[width=0.22\textwidth,clip=true,trim=0cm 0cm 0cm 0cm]{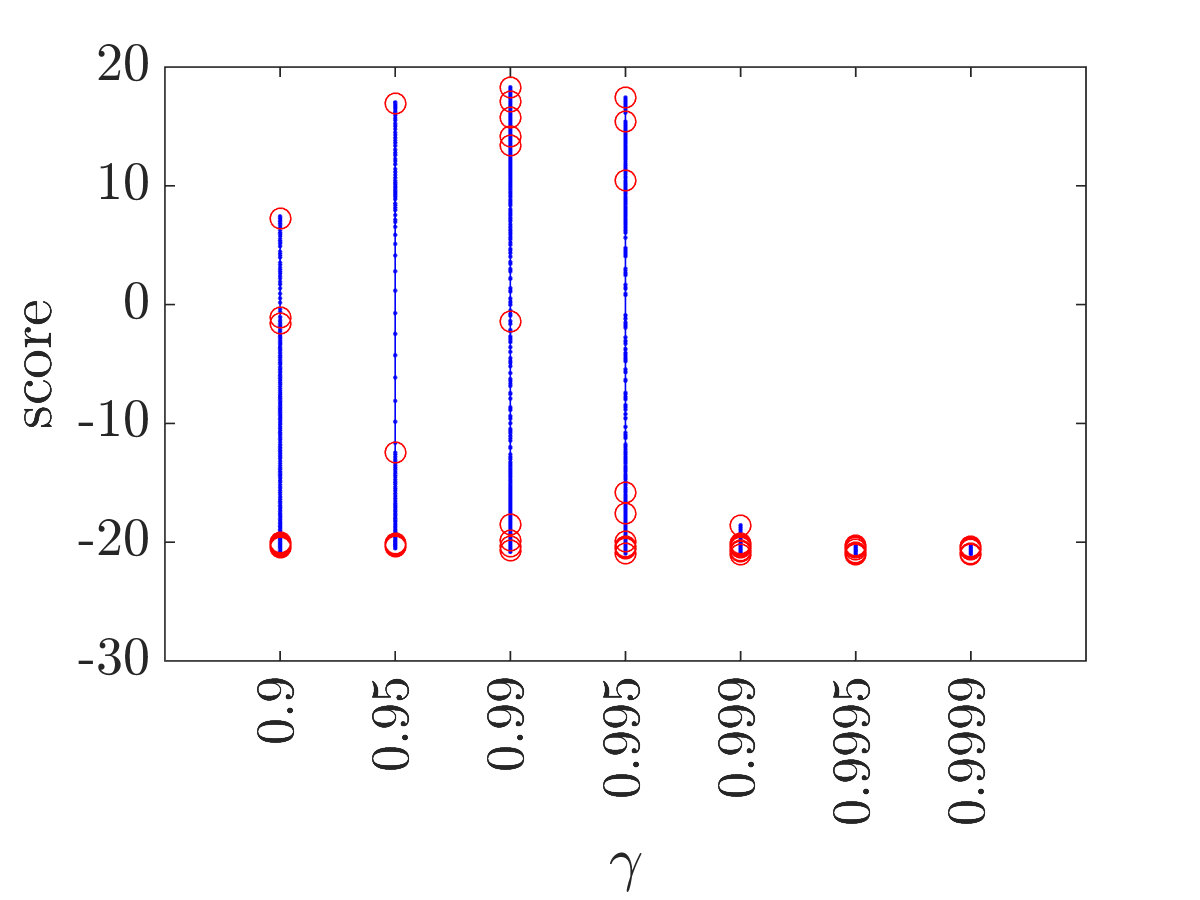}}}
    \subfloat{{\includegraphics[width=0.22\textwidth,clip=true,trim=0cm 0cm 0cm 0cm]{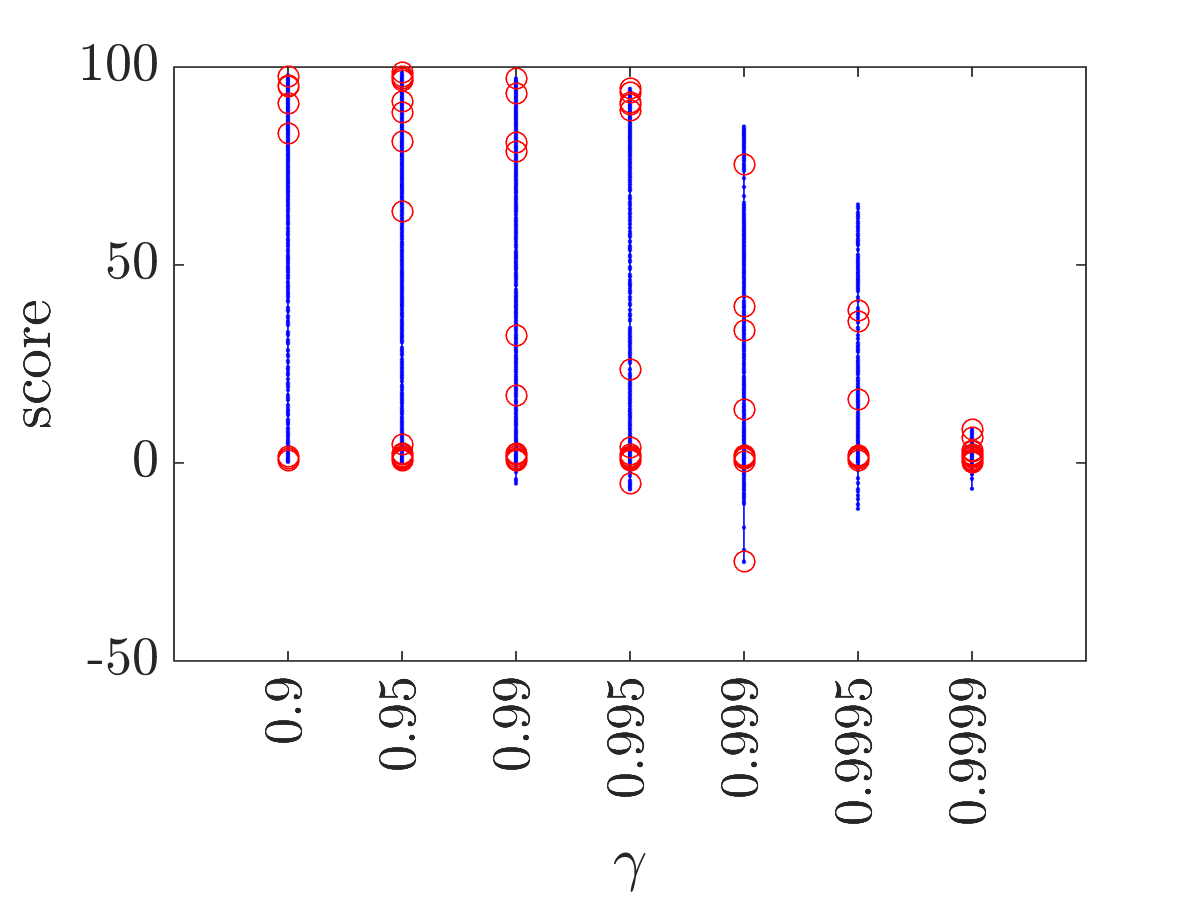}}}
    \subfloat{{\includegraphics[width=0.22\textwidth,clip=true,trim=0cm 0cm 0cm 0cm]{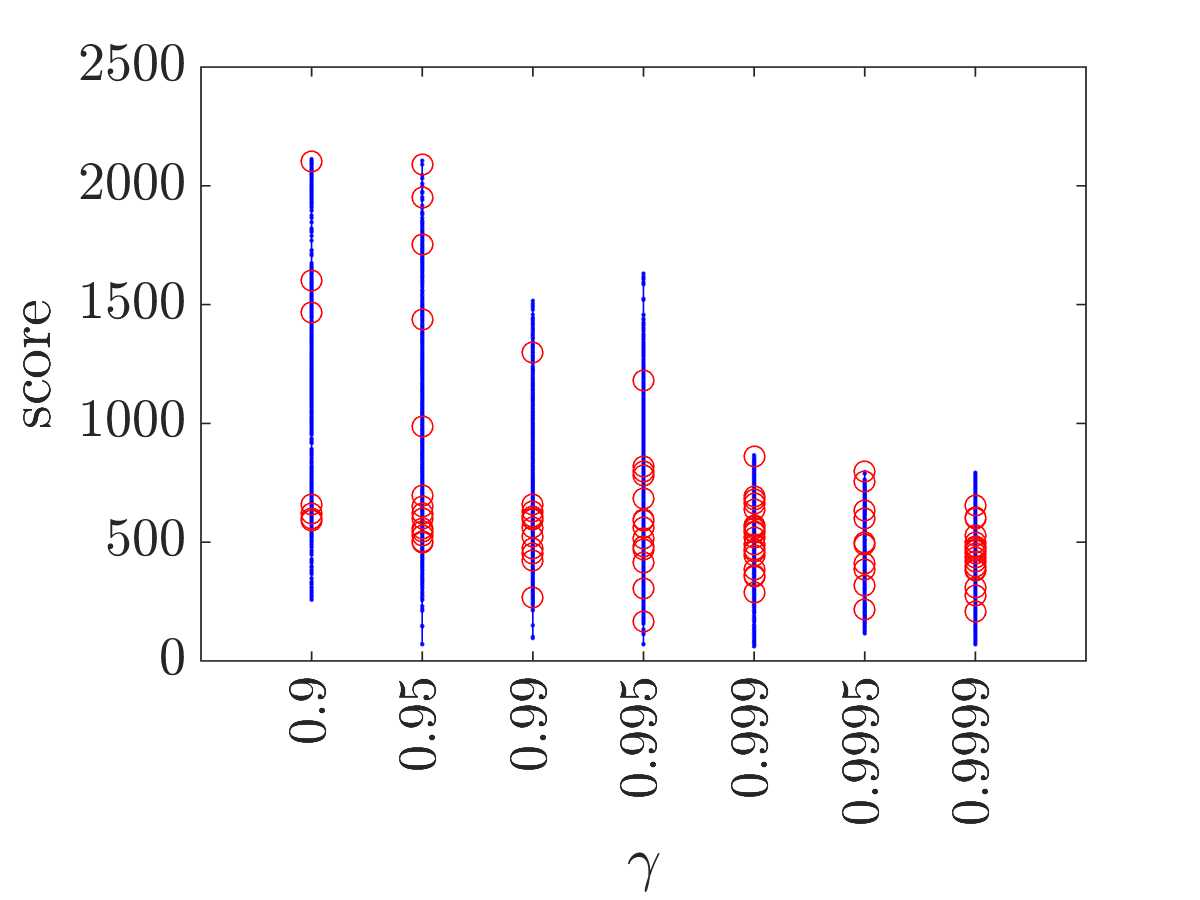}}}
    \subfloat{{\includegraphics[width=0.22\textwidth,clip=true,trim=0cm 0cm 0cm 0cm]{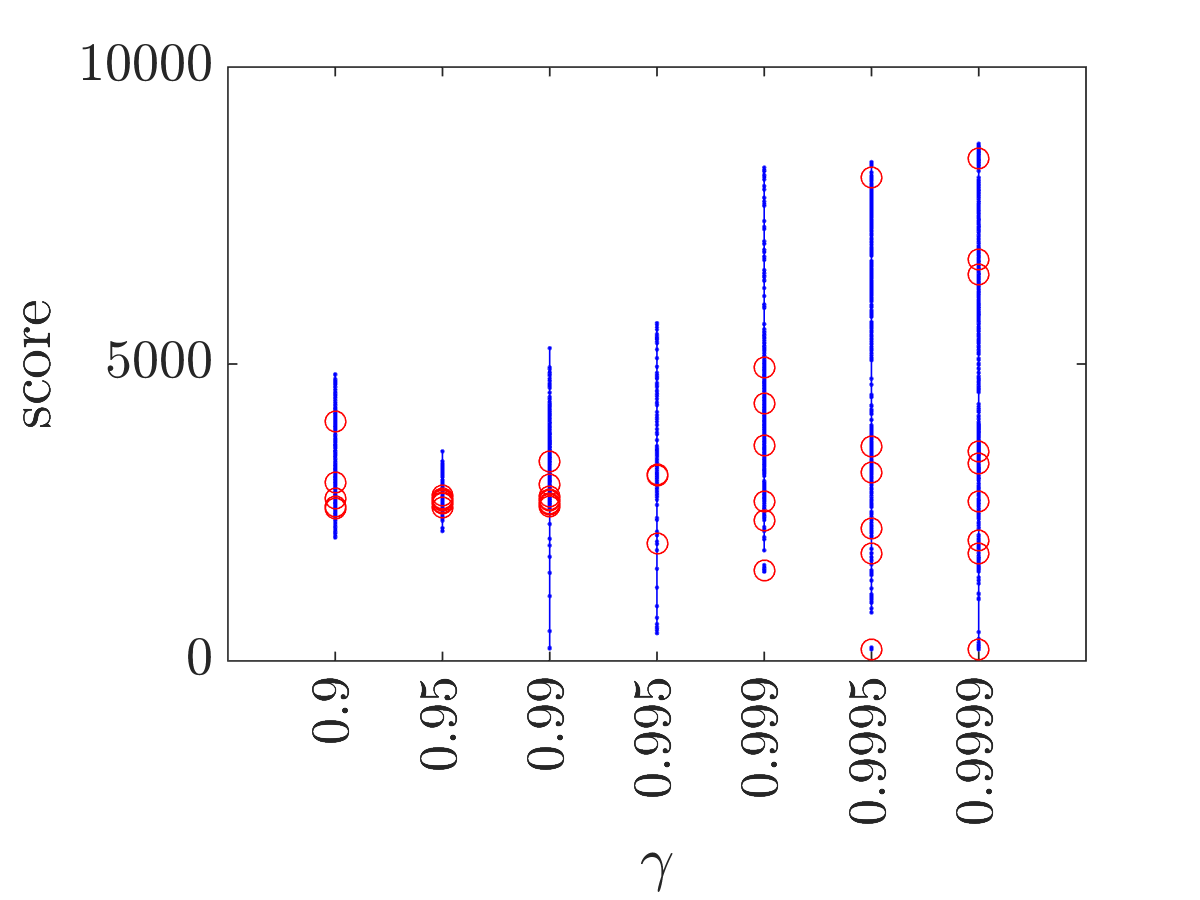}}}\\
    \vspace{-0.4cm}
    \setcounter{subfigure}{0}
    \subfloat[Pong]{{\includegraphics[width=0.22\textwidth,clip=true,trim=0cm 0cm 0cm 0.0cm]{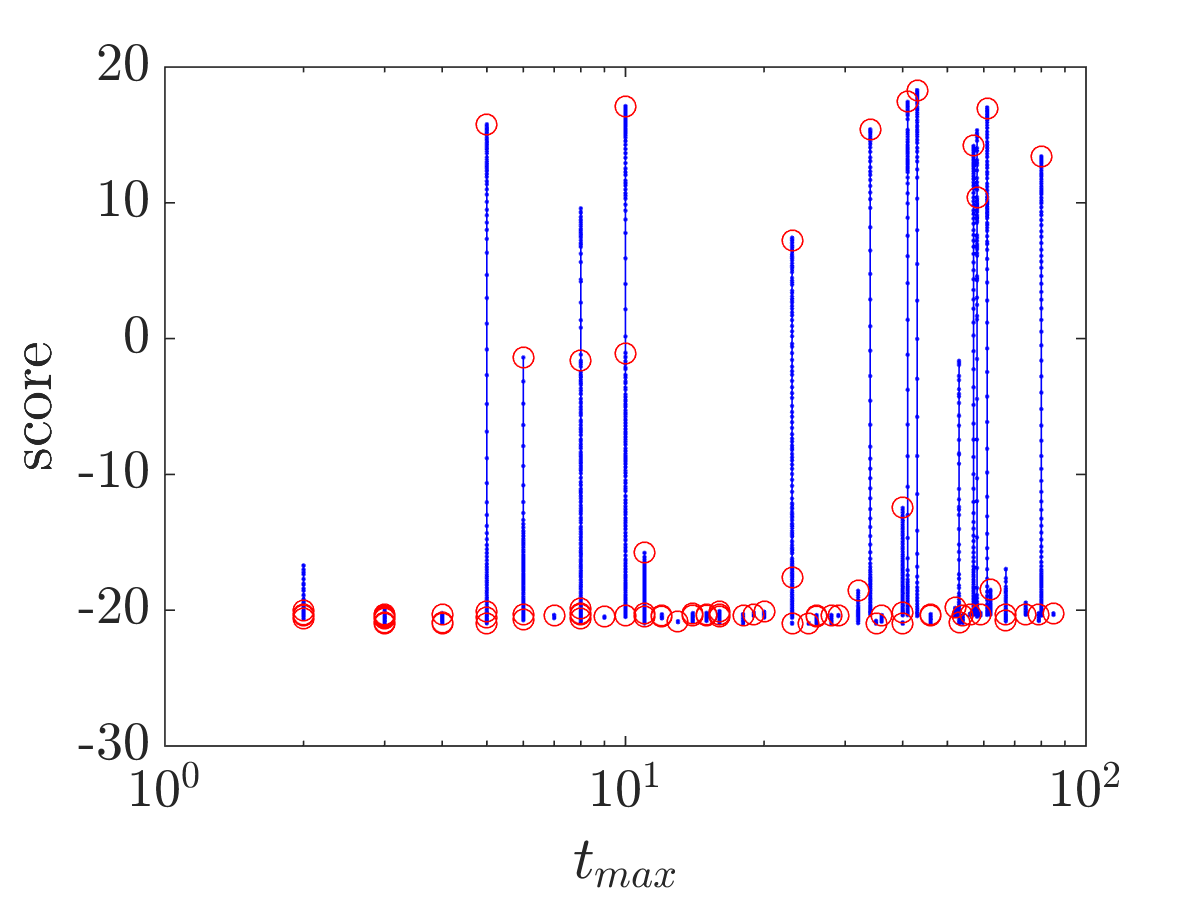}}}
    \subfloat[Boxing]{{\includegraphics[width=0.22\textwidth,clip=true,trim=0cm 0cm 0cm 0cm]{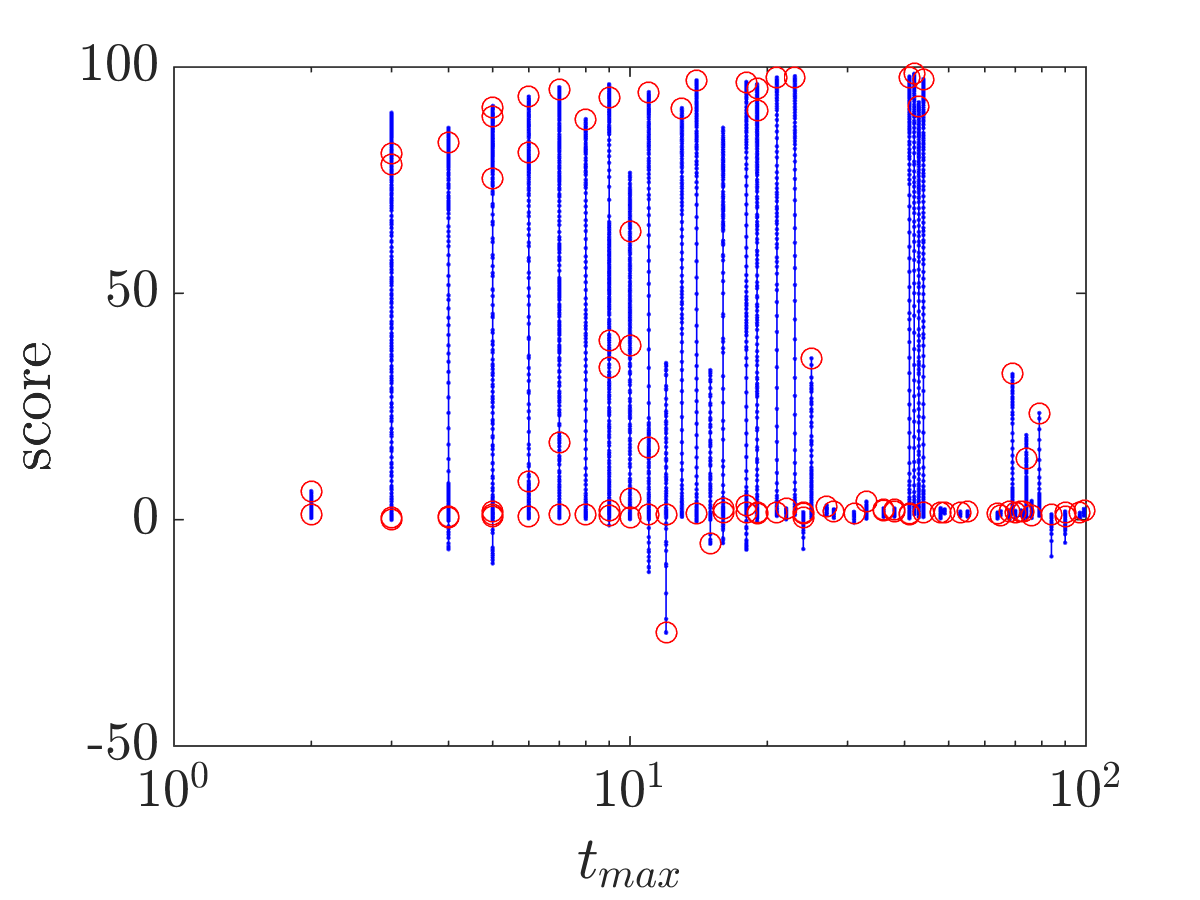}}}
    \subfloat[Ms Pacman]{{\includegraphics[width=0.22\textwidth,clip=true,trim=0cm 0cm 0cm 0cm]{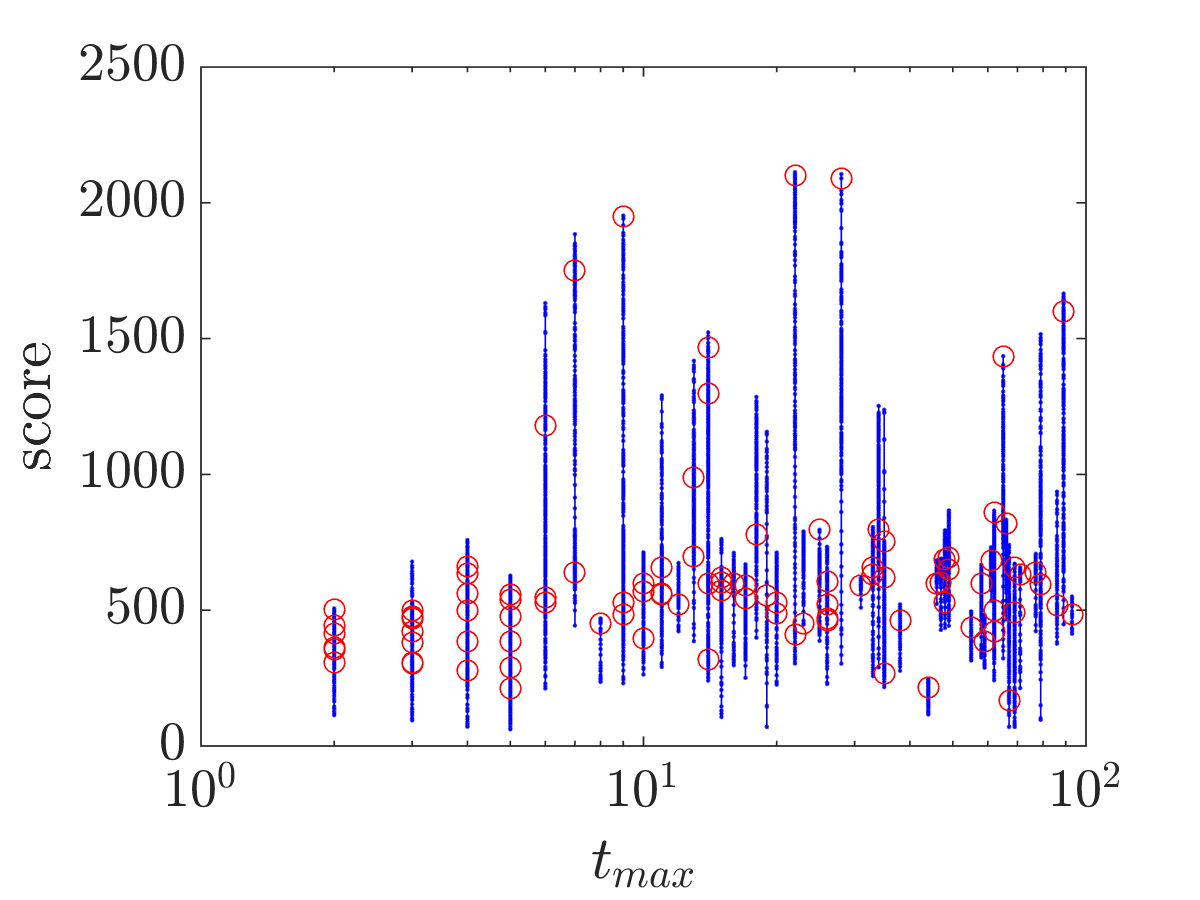}}}
    \subfloat[Centipede]{{\includegraphics[width=0.22\textwidth,clip=true,trim=0cm 0cm 0cm 0cm]{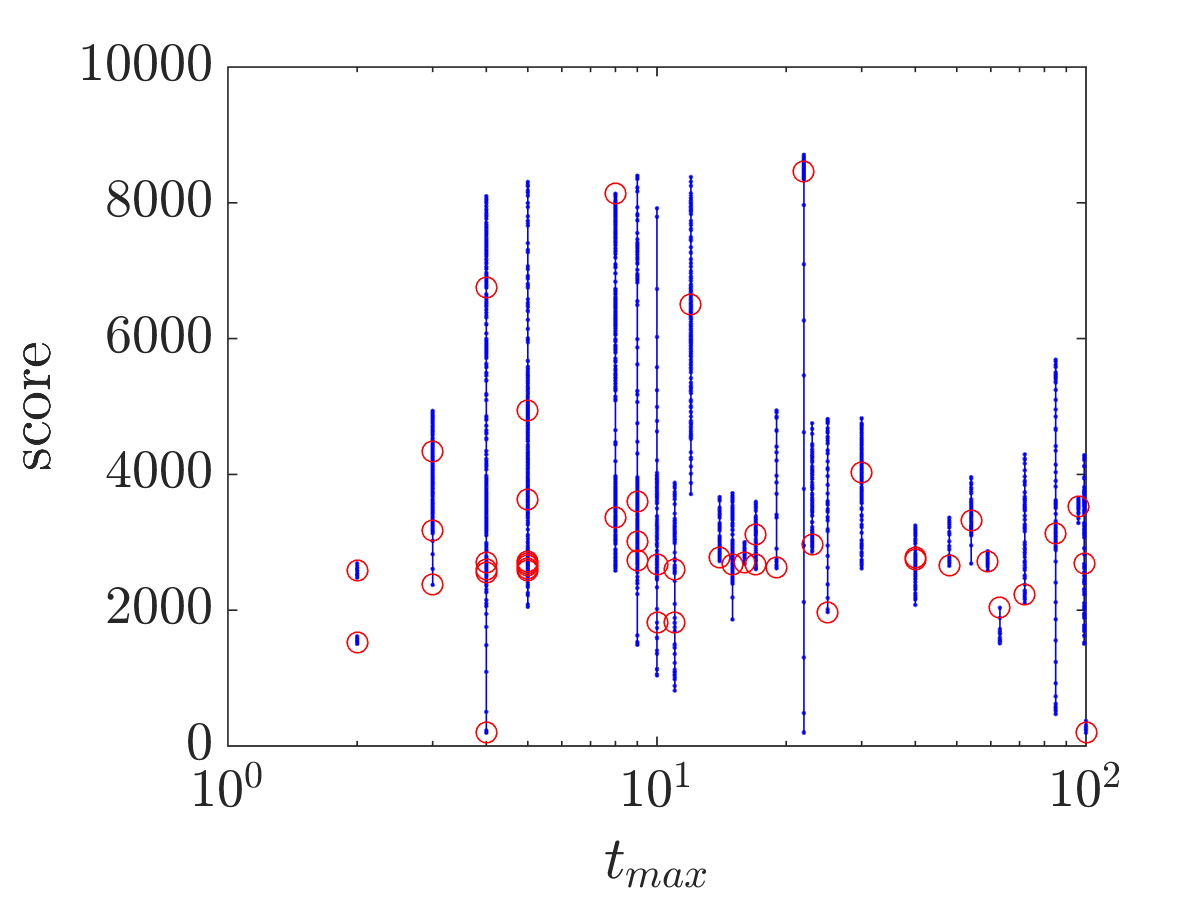}}}\\
    \caption{The final (red circles) and during training (blue lines) scores for 100 workers during metaoptimization through HyperTrick, as a function of the learning rate, discount $\gamma$, and $t_{max}$, for four Atari games learned with GA3C.
    }
    \label{fig:importanceOfHyperparameters}
\end{figure*}

Based on the results stored in the knowledge database in MagLev, we investigate the selection of the optimal hyperparameters for each game and how they affect the learned policy, for the experimental setup described in Section~\ref{subsec:experimental_setup}.
Fig.~\ref{fig:importanceOfHyperparameters} shows the distribution of the scores as a function of learning rate, $\gamma$, and $t_{max}$.
Since these scores are those reported by HyperTrick, not all the instances of GA3C run to completion.

The first row in Fig.~\ref{fig:importanceOfHyperparameters} highlights the importance of selecting a proper learning rate: for instance, in the case of Pong, a learning rate in the $[1.5\cdot10^{-4}, 3\cdot10^{-3}]$ interval is needed to solve the game; the situation is similar, although with different numerical intervals, for Boxing and Ms Pacman, whereas for Centipede a scattered set of learning rates generate the best scores.
As expected, agents learned with different learning rates do not show significant differences in the policy; the learning rate only affects the stability and rate of convergence towards the optimal policy.

The learning rate alone is clearly not sufficient to determine the success of RL training: the second row of Fig. \ref{fig:importanceOfHyperparameters} highlights the importance of the discount factor $\gamma$.
A common choice is to set $\gamma=0.99$ (\cite{Lil15,Mni16,Bab17}), but $\gamma$ values in a larger interval may be effectively employed for Pong and Boxing, whereas respectively smaller and larger $\gamma$ values potentially lead to better results for Ms-Pacman and Centipede.
Generally speaking, different games are learned at best for different intervals of $\gamma$ values.
This has been already noticed for instance in~\cite{Fra15}, where $\gamma$ is even modulated during training.
It can be explained noticing that the temporal dynamics of the rewards are different in each game: for instance, a reward is immediately generated when the adversary scores a goal in Pong; when one of the two players hits the other one in Boxing; or when a pill is eaten in Ms-Pacman; these examples justify the adoption of small $\gamma$ values for these games.
On the other hand, rewards in Centipede are delayed from the moment in which the player fires and the instant in which a target is hit, which may justify the preference for larger $\gamma$ values.

Quite remarkably, $\gamma$ does have a significant effect on the learned policy.
Agents that achieve similar scores on a game, but trained with different $\gamma$, adopt clearly different strategies.
For instance, short-sighted (small $\gamma$) agents in Pong mostly learn to not lose the game, moving somewhat erratically to catch the ball; these agents tend to engage in long rallies, waiting for the opponent to commit a mistake.
On the other hand, agents trained with a large $\gamma$ learn to hit the ball on the edge of the racket, to give it a spin and effectively score a goal some frames later.
In Boxing, short-sighted agents perform better as they kick the opponent as fast as possible, whereas looking for more long-term reward seems to merely give the opponent a chance to strike back.
In Centipede, all agents learn that firing as fast as possible is desirable.
High values of $\gamma$  encourage agents to hide towards the edge of the screen where aliens are less likely to attack, whereas slightly lower values of $\gamma$ force agents to stick to the center of the screen, where they are more exposed to attacks.
In Ms-Pacman, all the best agents are short-sighted ($\gamma < 0.9$); they learn to navigate the maze to eat close pills and escape from ghosts, but tend not to move towards far, isolated pills when few of them are left in the maze.
Overall, metaoptimization can effectively identify an optimal value for $\gamma$, but the learned policy is affected by this choice.
This interaction between one hyperparameter and the solution of the underneath optimization problem is easily justified in the context of RL, remembering that the discount factor $\gamma$ affects the definition of optimality of the learned policy; it is anyway a peculiar aspect of metaoptimization on RL problems, that has to be taken into consideration by future researchers working in this direction.

The last hyperparameter analyzed here is $t_{max}$, which is set by default to $t_{max} = 5$ in A3C~\cite{Mni16}.
The third row in Fig.~\ref{fig:importanceOfHyperparameters} suggests that higher values of $t_{max}$ (but not smaller ones) can indeed lead to convergence, although it's not evident whether an optimal interval for this hyperparameter can be identified.
Most likely, $t_{max}$ affects at the same time the computational cost of the learning procedure, by changing the size of the batch, and the noise level of the updates, by affecting the bias-variance trade-off in the estimate of the value function.
The relation between $t_{max}$ and the outcome of the RL procedure is consequently complex, and probably influenced by the other hyperparameters - thus metaoptimization helps to automatically identify the best value of $t_{max}$ in the absence of any other intuition.

The information stored by  MagLev in the central knowledge database can also be used to perform \emph{a posteriori} analyses that reveal more quantitative information about the RL procedure.
An example of this is reported in the Appendix, where we show how to train a regressor to estimate the relation between the hyperparameters and the final score achieved by the RL procedure, and consequently quantify the contribution of each hyperparameter to the success of the RL training procedure.

\section{Conclusion}

HyperTrick, the asynchronous metaoptimization algorithm proposed here, is particularly suitable for the case of distributed systems, when the selection of the hyperparameters affect the computational cost of the underneath experiments.
We demonstrate that HyperTrick allows effective metaoptimization for deep RL problems.
HyperTrick does not require any complex synchronization mechanisms or preemption management: it frees and reallocates computational resources more efficiently than algorithms based on the Successive Halving principle, like HyperBand.
When compared experimentally with those algorithms, HyperTrick achieves a higher occupancy of the nodes in the distributed system, completes the metaoptimization procedure in a shorter time, and finds the optimal solution earlier.
By adopting a stochastic process for the selection of the promising workers, HyperTrick gives early workers a higher chance to continue and increase the depth of their search, while late workers are discouraged.
HyperTrick achieves in this way a partial balance between breadth and depth search in metaoptimization.
A promising direction to achieve an even better balance is the integration of HyperTrick and Hyperband, where multiple instances of HyperTrick with different $N_p$ and $r$ may run in parallel.
Futhermore, the additional resources released by HyperTrick may be employed to further improve the metaoptimization process, for instance by the integration of evolutionary strategies, \emph{e.g.} by mixing the hyperparameters of fast learners, or reinitializing terminated agents with new sets of promising hyperparameters.
We leave these and other possible improvements for future investigation.
Our experiments finally highlight that, in the case of RL, hyperparameter selection and the learned policy can be connected, as in the case of the discounting factor $\gamma$ - this is an additional source of complexity that has to be taken into account by future researchers working in this field.





\bibliographystyle{sysml2019}
\bibliography{references}


\clearpage

\section{Appendix}
\label{sec:appendix}

\subsection{Metaoptimization with Static Allocation of the Workers to the Nodes}

For completeness, we report here simulations on the same metaoptimization toy problem introduced in Fig. \ref{fig:waveWithHyperTrick}, in the case of Successive Halving, with a static assignment of any worker to one node of the distributed system, and in the case of Grid Search (with no early stopping).

\begin{figure}[h]
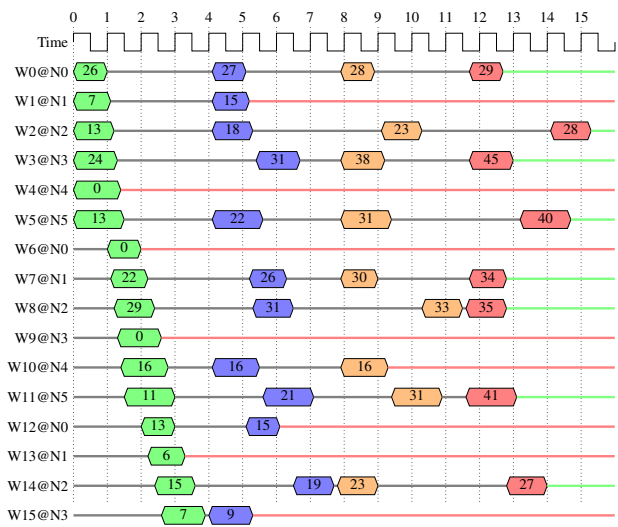

    \centering
    \resizebox{\linewidth}{!}{
\begin{wave}[Time]{16}{15}
\nextwave{W0@N0} \firstphase{26}{1.00} \pending{3.10} \secondphase{27}{1.00} \pending{2.80} \thirdphase{28}{1.00} \pending{2.80} \fourthphase{29}{1.00} \completed{3.30}
   \nextwave{W1@N1} \firstphase{7}{1.10} \pending{3.00} \secondphase{15}{1.10} \killed{10.80}
   \nextwave{W2@N2} \firstphase{13}{1.20} \pending{2.90} \secondphase{18}{1.20} \pending{3.80} \thirdphase{23}{1.20} \pending{3.80} \fourthphase{28}{1.20} \completed{0.70}
   \nextwave{W3@N3} \firstphase{24}{1.30} \pending{4.10} \secondphase{31}{1.30} \pending{1.20} \thirdphase{38}{1.30} \pending{2.50} \fourthphase{45}{1.30} \completed{3.00}
   \nextwave{W4@N4} \firstphase{0}{1.40} \killed{14.60}
   \nextwave{W5@N5} \firstphase{13}{1.50} \pending{2.60} \secondphase{22}{1.50} \pending{2.30} \thirdphase{31}{1.50} \pending{3.80} \fourthphase{40}{1.50} \completed{1.30}
   \nextwave{W6@N0} \pending{1.00} \firstphase{0}{1.00} \killed{14.00}
   \nextwave{W7@N1} \pending{1.10} \firstphase{22}{1.10} \pending{3.00} \secondphase{26}{1.10} \pending{1.60} \thirdphase{30}{1.10} \pending{2.70} \fourthphase{34}{1.10} \completed{3.20}
   \nextwave{W8@N2} \pending{1.20} \firstphase{29}{1.20} \pending{2.90} \secondphase{31}{1.20} \pending{3.80} \thirdphase{33}{1.20} \pending{0.10} \fourthphase{35}{1.20} \completed{3.20}
   \nextwave{W9@N3} \pending{1.30} \firstphase{0}{1.30} \killed{13.40}
   \nextwave{W10@N4} \pending{1.40} \firstphase{16}{1.40} \pending{1.30} \secondphase{16}{1.40} \pending{2.40} \thirdphase{16}{1.40} \killed{6.70}
   \nextwave{W11@N5} \pending{1.50} \firstphase{11}{1.50} \pending{2.60} \secondphase{21}{1.50} \pending{2.30} \thirdphase{31}{1.50} \pending{0.70} \fourthphase{41}{1.50} \completed{2.90}
   \nextwave{W12@N0} \pending{2.00} \firstphase{13}{1.00} \pending{2.10} \secondphase{15}{1.00} \killed{9.90}
   \nextwave{W13@N1} \pending{2.20} \firstphase{6}{1.10} \killed{12.70}
   \nextwave{W14@N2} \pending{2.40} \firstphase{15}{1.20} \pending{2.90} \secondphase{19}{1.20} \pending{0.10} \thirdphase{23}{1.20} \pending{3.80} \fourthphase{27}{1.20} \completed{2.00}
   \nextwave{W15@N3} \pending{2.60} \firstphase{7}{1.30} \pending{0.10} \secondphase{9}{1.30} \killed{10.70}
\end{wave}}
\caption{Metaoptimization with a variant of Successive Halving, which terminates 25\% of workers at the end of every phase, for the same toy problem in Fig. \ref{fig:waveWithHyperTrick}. Each worker is statically assigned to a single node for the entire process.
The process lasts for 15.3 units of time.}
\label{fig:waveWithSuccessiveHalvingFixedNodeAllocation}
\end{figure}

An implementation of Successive Halving with a static association between workers and nodes (Fig. \ref{fig:waveWithSuccessiveHalvingFixedNodeAllocation})  is possible, although a mechanism to manage preemption is needed even in this case, for two reasons: fast or early workers must stop and wait for all the other workers to complete a phase; in the case the number of nodes is smaller than the number of workers, the same node may have to run the same phase for more than one worker.
At least for the toy problem considered here, such implementation is extremely inefficient in terms of time consumption, as it takes 15.3 units of time (to be compared against 12.1 units of time for Successive Halving with dynamic allocation of the workers to the nodes in Fig.~\ref{fig:waveWithSuccessiveHalving}, and 10 units of time for HyperTrick in Fig. \ref{fig:waveWithHyperTrick}).

\begin{figure}[h]
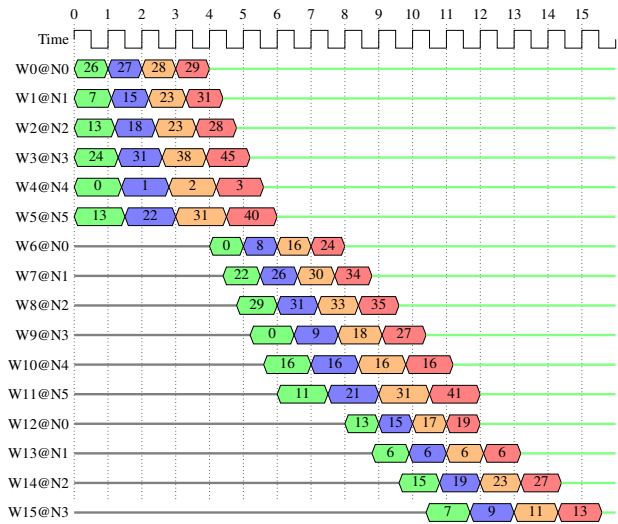

    \centering
    \resizebox{\linewidth}{!}{
\begin{wave}[Time]{16}{15}
\nextwave{W0@N0} \firstphase{26}{1.00} \secondphase{27}{1.00} \thirdphase{28}{1.00} \fourthphase{29}{1.00} \completed{12.00}
   \nextwave{W1@N1} \firstphase{7}{1.10} \secondphase{15}{1.10} \thirdphase{23}{1.10} \fourthphase{31}{1.10} \completed{11.60}
   \nextwave{W2@N2} \firstphase{13}{1.20} \secondphase{18}{1.20} \thirdphase{23}{1.20} \fourthphase{28}{1.20} \completed{11.20}
   \nextwave{W3@N3} \firstphase{24}{1.30} \secondphase{31}{1.30} \thirdphase{38}{1.30} \fourthphase{45}{1.30} \completed{10.80}
   \nextwave{W4@N4} \firstphase{0}{1.40} \secondphase{1}{1.40} \thirdphase{2}{1.40} \fourthphase{3}{1.40} \completed{10.40}
   \nextwave{W5@N5} \firstphase{13}{1.50} \secondphase{22}{1.50} \thirdphase{31}{1.50} \fourthphase{40}{1.50} \completed{10.00}
   \nextwave{W6@N0} \pending{4.00} \firstphase{0}{1.00} \secondphase{8}{1.00} \thirdphase{16}{1.00} \fourthphase{24}{1.00} \completed{8.00}
   \nextwave{W7@N1} \pending{4.40} \firstphase{22}{1.10} \secondphase{26}{1.10} \thirdphase{30}{1.10} \fourthphase{34}{1.10} \completed{7.20}
   \nextwave{W8@N2} \pending{4.80} \firstphase{29}{1.20} \secondphase{31}{1.20} \thirdphase{33}{1.20} \fourthphase{35}{1.20} \completed{6.40}
   \nextwave{W9@N3} \pending{5.20} \firstphase{0}{1.30} \secondphase{9}{1.30} \thirdphase{18}{1.30} \fourthphase{27}{1.30} \completed{5.60}
   \nextwave{W10@N4} \pending{5.60} \firstphase{16}{1.40} \secondphase{16}{1.40} \thirdphase{16}{1.40} \fourthphase{16}{1.40} \completed{4.80}
   \nextwave{W11@N5} \pending{6.00} \firstphase{11}{1.50} \secondphase{21}{1.50} \thirdphase{31}{1.50} \fourthphase{41}{1.50} \completed{4.00}
   \nextwave{W12@N0} \pending{8.00} \firstphase{13}{1.00} \secondphase{15}{1.00} \thirdphase{17}{1.00} \fourthphase{19}{1.00} \completed{4.00}
   \nextwave{W13@N1} \pending{8.80} \firstphase{6}{1.10} \secondphase{6}{1.10} \thirdphase{6}{1.10} \fourthphase{6}{1.10} \completed{2.80}
   \nextwave{W14@N2} \pending{9.60} \firstphase{15}{1.20} \secondphase{19}{1.20} \thirdphase{23}{1.20} \fourthphase{27}{1.20} \completed{1.60}
   \nextwave{W15@N3} \pending{10.40} \firstphase{7}{1.30} \secondphase{9}{1.30} \thirdphase{11}{1.30} \fourthphase{13}{1.30} \completed{0.40}
\end{wave}}
\caption{Metaoptimization with Grid Search (no early stopping) on the same toy problem in Fig. \ref{fig:waveWithHyperTrick}. No preemption mechanism is needed in this case. The process lasts 15.6 units of time.}
\label{fig:waveWithoutHyperTrick}
\end{figure}

If preemption cannot be implemented, the metaoptimization scheme boils down to a full Grid Search, shown in Fig.~\ref{fig:waveWithoutHyperTrick}.
On the toy problem considered here, this is the slowest metaoptimization scheme (15.6 units of time).
Since all the workers run the underneath optimization experiments to the end, the worker completion rate is in this case equal to $\alpha = 100\%$.

\subsection{Estimate the Importance of the Hyperparameters}

The information stored by  MagLev in the central knowledge database can  be used to perform \emph{a posteriori} analyses that reveal quantitative information about the RL procedure.
An example is illustrated in the following, where we estimate the importance of the learning rate, $\gamma$ and $t_{max}$ to determine the final score of a game.

\begin{table}[h]
\centering
  \begin{tabular}{|c|c|c|c|}
    \hline
    Game & learning rate & $\gamma$ & $t_{max}$ \\ \hline
    Boxing & 54\% & 24\% & 22\% \\
    Centipede & 34\% & 35\% & 31\% \\
    Ms Pacman & 43\% & 37\% & 20\% \\
    Pong & 55\% & 31\% & 14\% \\
    \hline
  \end{tabular}
\caption{Importance of the learning rate, $\gamma$ and $t_{max}$ for every game, as estimated by a Random Forest regressor trained to map a hyperparameter configuration to a score.}
\label{table:featureimportances}
\end{table}

Although learning an accurate mapping function between the hyperparameters and the final score is complex, we show that even an approximated function can provide valuable insights about the role played by each hyperparameter.
To show this, we employ a Random Forest regressor trained to map a hyperparameter configuration to the score achieved in HyperTrick.
Notice that this score is not necessarily the final score after completing all the phases, as a worker can be terminated early.
For each game we use Scikit Learn~\cite{scikit-learn} to train 100 Random Forest regressors using various configurations of the Random Forest parameters; we use then 10-fold cross validation to identify the best, non overfitting, regressor.
The feature importances for each game, extracted through the Scikit API, are reported in Table~\ref{table:featureimportances}.

Centipede arguably features the noisiest learning curves and unsurprisingly its regressor gives identical
importance to all the hyperparameters.
Conversely, for Pong $\gamma$ and the learning rate dominate, and $t_{max}$ appears to be less important.
These results are in line with the intuition one can build from Fig. \ref{fig:hyperparametersSelection} and the observations reported in the main paper; they confirm the general small influence of $t_{max}$ on the final score, and the importance of a proper selection of the learning rate.

\clearpage

\end{document}